\PassOptionsToPackage{table}{xcolor}
\documentclass[runningheads]{llncs}

\usepackage{eccv}

\usepackage{eccvabbrv}

\usepackage{graphicx}
\usepackage{booktabs}
\usepackage{multirow}
\usepackage{pifont}
\usepackage{adjustbox}
\usepackage{makecell}
\usepackage[breaklinks,colorlinks,citecolor=eccvblue,urlcolor=black,linkcolor=black]{hyperref}
\usepackage[accsupp]{axessibility}

\definecolor{softgray}{RGB}{242, 242, 242}
\definecolor{softblue}{RGB}{208, 231, 255}
\definecolor{skyblue}{RGB}{135, 206, 235}
\definecolor{lightblue}{RGB}{173, 216, 230}
\definecolor{lightyellow}{RGB}{255, 255, 224}

\begin{document}

\title{UniREditBench: A Unified Reasoning-based Image Editing Benchmark}
\titlerunning{UniREditBench: A Unified Reasoning-based Image Editing Benchmark}

\author{
Feng Han\inst{1,2}$^{*}$ \and
Yibin Wang\inst{1,2}$^{*}$ \and
Chenglin Li\inst{2,3} \and
Zheming Liang\inst{2} \and \\
Dianyi Wang\inst{1,2} \and
Yang Jiao\inst{1} \and
Zhipeng Wei\inst{4} \and
Chao Gong\inst{1} \and \\
Cheng Jin\inst{1,2} \and
Jiaqi Wang\inst{2,5}$^{\dagger}$
}

{
  \renewcommand{\thefootnote}{\fnsymbol{footnote}}
  \footnotetext[1]{Equal contribution. \textsuperscript{\dag} Corresponding author.
  }
}

\authorrunning{F. Han et al.}

\institute{
\textsuperscript{\rm 1} Fudan University \ 
\textsuperscript{\rm 2} Shanghai Innovation Institute \
\textsuperscript{\rm 3} Zhejiang University \\ 
\textsuperscript{\rm 4} UC Berkeley \ 
\textsuperscript{\rm 5} JD.COM \\
\textbf{Project Page}: \url{https://maplebb.github.io/UniREditBench/}
}

\maketitle

\begin{figure*}[t]
    \centering
    \includegraphics[width=0.95\linewidth]{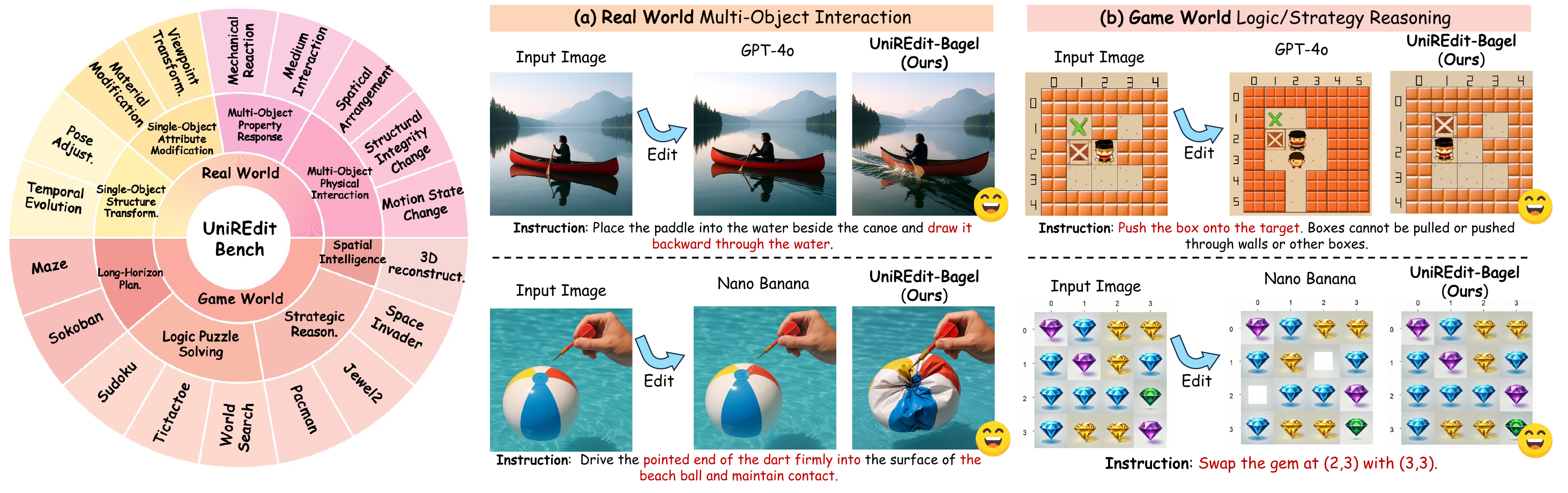}
    \caption{UniREditBench covers both real-world and game-world reasoning scenarios across 8 primary dimensions and 18 sub-dimensions. We provide editing cases of (a) real-world multi-object interaction, and (b) game-world logical/strategy reasoning.}
    \label{fig:teaser}
    \vspace{-0.5cm}
\end{figure*}

\begin{abstract}
Recent advances in multimodal generative models have driven substantial improvements in image editing. However, current generative models still struggle with handling diverse and complex image editing tasks that require implicit reasoning, underscoring the need for a comprehensive benchmark to systematically assess their performance across various reasoning scenarios. Existing benchmarks primarily focus on single-object attribute transformation in realistic scenarios, which, while effective, encounter two key challenges: \textbf{(1)} they largely overlook multi-object interactions as well as game-world scenarios that involve human-defined rules, which are common in real-life applications; \textbf{(2)} they only rely on textual references to evaluate the generated images, potentially leading to systematic misjudgments, especially in complex reasoning scenarios. To this end, this work proposes \textbf{UniREditBench}, a unified benchmark for reasoning-based image editing evaluation. It comprises 2,700 meticulously curated samples, covering both real- and game-world scenarios across 8 primary dimensions and 18 sub-dimensions. To improve evaluation reliability, we introduce multimodal dual-reference evaluation, providing both textual and ground-truth image references for each sample assessment. Furthermore, we design an automated multi-scenario data synthesis pipeline and construct \textbf{UniREdit-Data-100K}, a large-scale synthetic dataset with high-quality chain-of-thought (CoT) reasoning annotations. We fine-tune Bagel on this dataset and develop \textbf{UniREdit-Bagel}, demonstrating substantial improvements in both in-domain and out-of-distribution settings. Through thorough benchmarking of both open-source and closed-source image editing models, we reveal their strengths and weaknesses across various aspects.
\end{abstract}
\section{Introduction}
\label{sec:intro}

\definecolor{navy_blue}{RGB}{0, 47, 167}
\definecolor{richCrimson}{RGB}{153, 0, 51} 
\definecolor{red_1}{RGB}{241, 185, 184}
\definecolor{blue_1}{RGB}{201, 224, 251}
\definecolor{UniPink}{RGB}{255,235,242} 
\definecolor{UniRow}{RGB}{235,245,255} 

Recent advances in multimodal generative models have led to remarkable improvements in instruction-conditioned image editing. Generative models~\cite{wu2025omnigen2, yu2025anyedit, zhang2023magicbrush, brooks2023instructpix2pix,wang2024primecomposer,wang2025dreamtext,xin2025lumina}, including Step1X-Edit~\cite{liu2025step1x}, FLUX-Kontext~\cite{flux_kontext}, Bagel~\cite{deng2025bagel}, Nano Banana~\cite{Gemini-2.5-Flash-Image}, and GPT-4o~\cite{hurst2024gpt}, have demonstrated a powerful ability to understand diverse textual instructions and generate semantically consistent image edits. In parallel, reinforcement learning-based training strategies~\cite{xiao2025mindomni,li2025uniworld,wang2025pref,wang2024lift} are continuously advancing, further enhancing the capabilities of image editing models. With these rapid developments, the need for a more comprehensive benchmark to evaluate model editing capabilities across different aspects has become increasingly essential. Early benchmarks~\cite{liu2025step1x, ye2025imgedit} focus on local details or global stylistic changes, e.g., style transfer, color alteration, and object removal. However, they fail to cover editing tasks that require models to perform implicit reasoning~\cite{zhang2025r, he2025reasoning}, which are commonly used in real-life applications. As illustrated in Fig.\textcolor{red}{~\ref{fig:teaser}}, when editing instructions involving real-world or human-defined game rules, current models often generate results that lack physical plausibility. To this end, recent efforts have introduced reasoning-aware evaluation across temporal, spatial, and logical dimensions~\cite{zhao2025envisioning}, and proposed a knowledge-grounded taxonomy assessing factual, conceptual, and procedural knowledge types~\cite{wu2025kris}.

\begin{figure}[t] 
    \centering
    \includegraphics[width=1.04\linewidth]{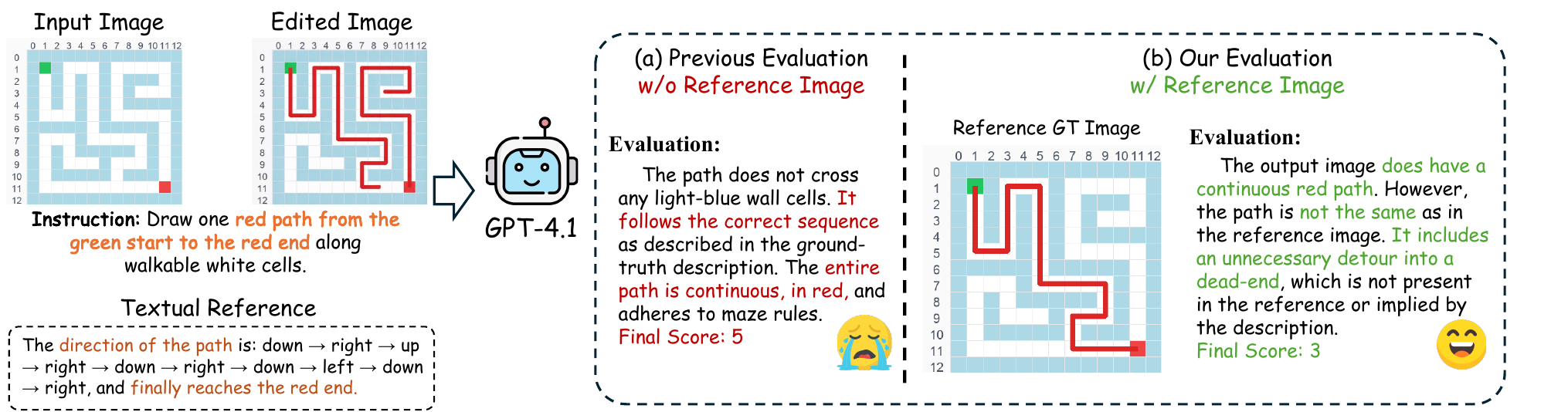} 
    \caption{Image editing evaluation comparison. Current text-reference-only evaluation potentially leads to misjudgments, while our dual-reference evaluation results in more reliable assessments.}
    \label{fig:eval_compare}
    \vspace{-0.6cm}
\end{figure}

Despite their effectiveness, these benchmarks still face two significant challenges: \textbf{(1)} they primarily focus on single-object attribute changes in realistic scenarios, neglecting multi-object interactions and game-world scenarios involving human-defined rules (see Tab. \textcolor{red}{\ref{tab:benchmark_compare}}). This narrow scope restricts their ability to evaluate how effectively models generalize across a wider range of complex reasoning contexts. Additionally, \textbf{(2)} they mainly rely on textual references to evaluate the generated images~\cite{zhao2025envisioning, wu2025kris}, which may lead to systematic misjudgments, especially in complex reasoning-based editing scenarios (see Fig.\textcolor{red}{~\ref{fig:eval_compare}}).

In this work, we posit that: 
\textbf{(1)} While current models exhibit proficiency in perceptual instruction following and simple reasoning editing settings (e.g., \textit{Transform an intact apple to a bitten one}), they still struggle with complex reasoning-based image editing that necessitates the comprehension of multi-object interaction characteristics (e.g., \textit{Draw the paddle backward through the water}) as well as logical constraints of puzzle and game scenarios (e.g., \textit{Control the player and push the box to the target}), as illustrated in Fig.\textcolor{red}{~\ref{fig:teaser}}.
\textbf{(2)} Relying on textual references in evaluating complex reasoning-based image editing tasks often leads to unreliable judgments. As shown in Fig. \textcolor{red}{~\ref{fig:eval_compare}} (a), the text-reference-only evaluator assigns an inflated score even when the edited image introduces an additional faulty path. Therefore, we intuitively believe that incorporating a ground-truth (GT) image as an additional visual reference can enable more precise evaluation. 

\begin{table*}[t]
    \centering
    \caption{Comparison of reasoning-based image editing benchmarks. ``S-Obj'' and ``M-Obj'' denote single-object and multi-object editing, respectively. UniREditBench stands out for its dual-reference evaluation protocol as well as broader coverage of real- and game-world scenarios.
    See Appendix for Detailed analysis.}
    \label{tab:benchmark_compare}
    \begin{adjustbox}{max width=\textwidth}{
    \renewcommand\arraystretch{1.05}
    \begin{tabular}{@{}l*{19}{c}@{}}
    \toprule
    \multirow{3}{*}{Benchmark} &
    \multirow{3}{*}{Size} &
    \multicolumn{3}{c}{\textbf{$\blacktriangledown$ Evaluation Validity}} &
    \multicolumn{8}{c}{\textbf{$\blacktriangledown$ Real World Scenario}} &
    \multicolumn{7}{c}{\textbf{$\blacktriangledown$ Game World Scenario}} \\
    \cmidrule(lr){3-5} \cmidrule(lr){6-13} \cmidrule(lr){14-20}
     &  &
     \makecell{Reference\\Image} &
     \makecell{Eval Ref.\\Modality\\(Text/Image)} &
     \makecell{Unified\\Dual-ref\\(Text+Image)} &
     \makecell{Attribute\\(S-Obj)} &
     \makecell{Temporal\\(S-Obj)} &
     \makecell{Pose\\(S-Obj)} &
     \makecell{Structure\\(M-Obj)} &
     \makecell{Spatial\\(M-Obj)} &
     \makecell{Motion\\(M-Obj)} &
     \makecell{Mechanic\\(M-Obj)} &
     \makecell{Medium\\(M-Obj)} &
     \makecell{Logical} &
     \makecell{Long-plan} &
     \makecell{Strategic} &
     \makecell{Spatial} &
     \makecell{Explicit\\Rules} &
     \makecell{Action\\Control} &
     \makecell{Intra-game\\Multi-task} \\
    \midrule

    SmartEdit~\cite{huang2024smartedit} &
    219 &
    \makecell{All\\(219/219)} &
    \makecell{Image-only} &
     &
    \ding{52} &  &  &  & \ding{52} &  &  &  &
     &  &  &  &
     &  &  \\

    RISE~\cite{zhao2025envisioning} &
    360 &
    \makecell{Partial\\(70/360)} &
    \makecell{Text-major} &
     &
    \ding{52} & \ding{52} &  &  & \ding{52} &  &  &  &
    \ding{52} & \ding{52} &  &  &
     &  &  \\

    KRIS~\cite{wu2025kris} &
    1,267 &
    \makecell{Partial\\(50/1,267)} &
    \makecell{Text-major} &
     &
    \ding{52} & \ding{52} &  &  & \ding{52} &  &  & \ding{52} &
    \ding{52} & \ding{52} &  &  &
     &  &  \\

    \midrule
    \rowcolor[HTML]{E2F4E3}
    \textbf{UniREditBench} &
    \textbf{2,700} &
    \parbox[c]{2.1cm}{\centering\textbf{All}\\\textbf{(2,700/2,700)}} &
    \textbf{Dual} &
    \ding{52} &
    \ding{52} & \ding{52} & \ding{52} & \ding{52} & \ding{52} & \ding{52} & \ding{52} & \ding{52} &
    \ding{52} & \ding{52} & \ding{52} & \ding{52} &
    \ding{52} & \ding{52} & \ding{52} \\
    \bottomrule
    \end{tabular}}
    \end{adjustbox}
    \vspace{-0.2cm}
\end{table*}

\begin{figure}[t]
    \centering
    \includegraphics[width=0.9\linewidth]{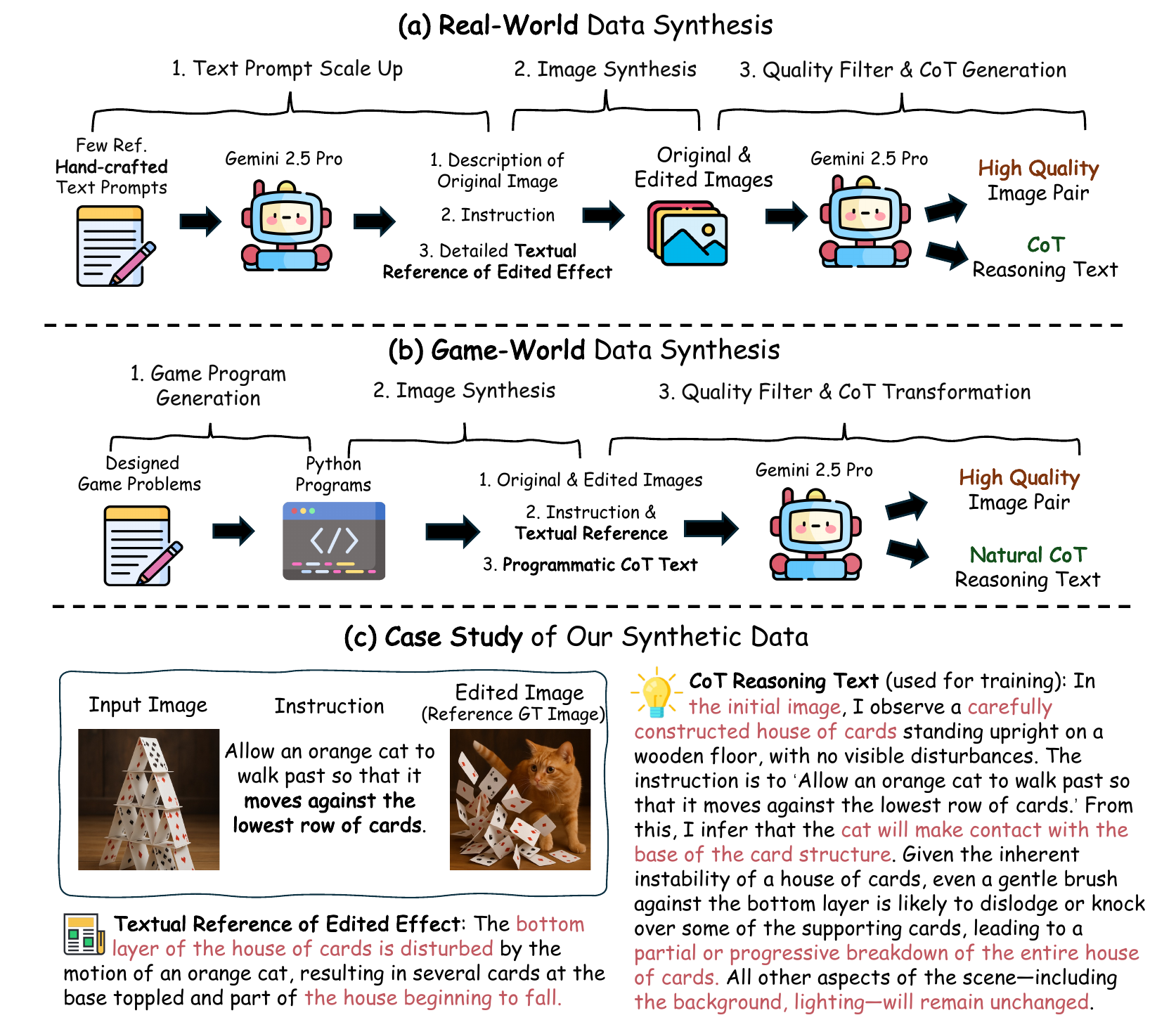}
    \caption{Multi-scenario data synthesis pipeline. (a) Real-world data synthesis pipeline; (b) Game-world data synthesis pipeline; and (c) Case study of our synthesized data.}
    \label{fig:data_construction}
    \vspace{-0.5cm}
\end{figure}

To this end, this work proposes \textbf{UniREditBench}, a unified benchmark for reasoning-based image editing assessment with broader evaluation dimension coverage and robust evaluation pipeline. Specifically, \textbf{(1)} we adopt a scenario-to-category hierarchical dimension design, covering diverse reasoning types in both real-world and game-world scenarios (shown in Fig.\textcolor{red}{~\ref{fig:teaser}}): it includes 2,700 carefully curated samples organized across 8 primary dimensions and 18 sub-categories, e.g., \textit{multi-object interaction} in real world, and \textit{long-horizon game planning} in game world. Meanwhile, \textbf{(2)} as illustrated in Fig.\textcolor{red}{~\ref{fig:eval_compare}}, in contrast to existing work that relies solely on textual references for evaluation, we introduce additional reference GT images to facilitate direct visual comparison with the generated image. By utilizing the visual cues provided by the reference image, the evaluator is able to more accurately and reliably assess the alignment of the generated image with the given instruction, as shown in Fig.\textcolor{red}{~\ref{fig:eval_compare}} (b). Furthermore, to ensure the diversity and reliability of samples in this benchmark, we design a \textbf{multi-scenario data synthesis pipeline}. Specifically, as shown in Fig.\textcolor{red}{~\ref{fig:data_construction}}, \textbf{(a)} For \textit{real-world} scenarios, we first handcraft a few reference text prompts, including the original image description, the editing instruction, and the textual reference of edited effect. These prompts are then scaled up using the VLM. Finally, all resulting textual descriptions are directly used to generate pairs of original and edited image. \textbf{(b)} For \textit{game-world} scenarios, we first design diverse game problems, and then use Python programs to generate image pairs, instructions, and textual references of edited effects, ensuring both logical and visual correctness in these rule-intensive scenarios~\cite{li2024vcbench, tong2025code2logic}.
Ultimately, all data samples in UniREditBench undergo VLM-based filtering and human inspection to ensure their reliability and accuracy.

Based on our data synthesis pipeline, we also propose \textbf{UniREdit-Data-100K}, a comprehensive reasoning-based image editing dataset with high-quality chain-of-thought (CoT) reasoning annotations, consisting of detailed, step-by-step reasoning traces generated using VLM, as shown in Fig.\textcolor{red}{~\ref{fig:data_construction}}.  To validate its reliability and effectiveness, we fine-tune the Bagel~\cite{deng2025bagel} on this dataset, resulting in \textbf{UniREdit-Bagel}.
Experimental results demonstrate that the fine-tuned model achieves substantial improvements on both UniREditBench and other out-of-distribution benchmarks~\cite{zhao2025envisioning, wu2025kris}. Additionally, through comprehensive evaluation of both open- and closed-source editing models on our UniREditBench, we reveal their strengths and weaknesses across diverse reasoning-based scenarios.

\textbf{Contributions:} (1) We introduce {UniREditBench}, a unified benchmark for reasoning-based image editing that covers both real-world and game-world scenarios across 8 primary dimensions and 18 sub-dimensions, augmented with reference GT images to enable robust evaluation;
(2) We design a {multi-scenario data synthesis pipeline} and develop {UniREdit-Data-100K}, a large-scale synthetic reasoning-based image editing dataset that includes high-quality CoT reasoning annotations. By fine-tuning the Bagel on this dataset, we develop UniREdit-Bagel and achieve substantial improvements, validating the effectiveness and reliability of our dataset;
(3) Through comprehensive benchmarking of both open- and closed-source models, we systematically identify their strengths and weaknesses across diverse reasoning-based editing scenarios, offering valuable insights for advancing future models.

\section{Related Work}
\noindent\textbf{Instruction-based Image Editing.}
Instruction-based image editing models aim to bridge semantic understanding of instructions with accurate visual manipulation. Traditional methods perform editing by altering the diffusion trajectory without requiring additional training, including partial denoising from intermediate SDE steps~\cite{meng2021sdedit}, cross-attention control~\cite{hertz2022prompt,wang2023enhancing}, mask-guided blending~\cite{wang2025dreamtext,avrahami2022blended,wang2024primecomposer}, CLIP- or diffusion-guided manipulation~\cite{kim2022diffusionclip,gong2026freeinpaint}, and latent inversion for fidelity preservation \cite{wang2024magicface,kawar2023imagic}. 
Besides, several studies employ visual-language models (VLMs) to provide prompts, spatial priors, or synthetic supervision to guide a generative editing model~\cite{brooks2023instructpix2pix, zhang2023magicbrush, shen2024imagpose, zhang2024hive, he2024freeedit, han2025controlthinker, shen2025imagdressing}.
Recent unified frameworks aim to use a single model for both image understanding and editing in a complementary direction~\cite{xin2025lumina, wu2025omnigen2, wang2024emu3}. For instance, Bagel~\cite{deng2025bagel} features a \emph{think} mode that produces reasoning text prior to editing to enhance instruction fidelity and consistency. While effective, current methods still face challenges with complex reasoning-based editing, underscoring the need for comprehensive benchmarks to assess their performance across various reasoning scenarios.

\begin{figure*}[t] 
    \centering
    \includegraphics[width=0.87\textwidth]{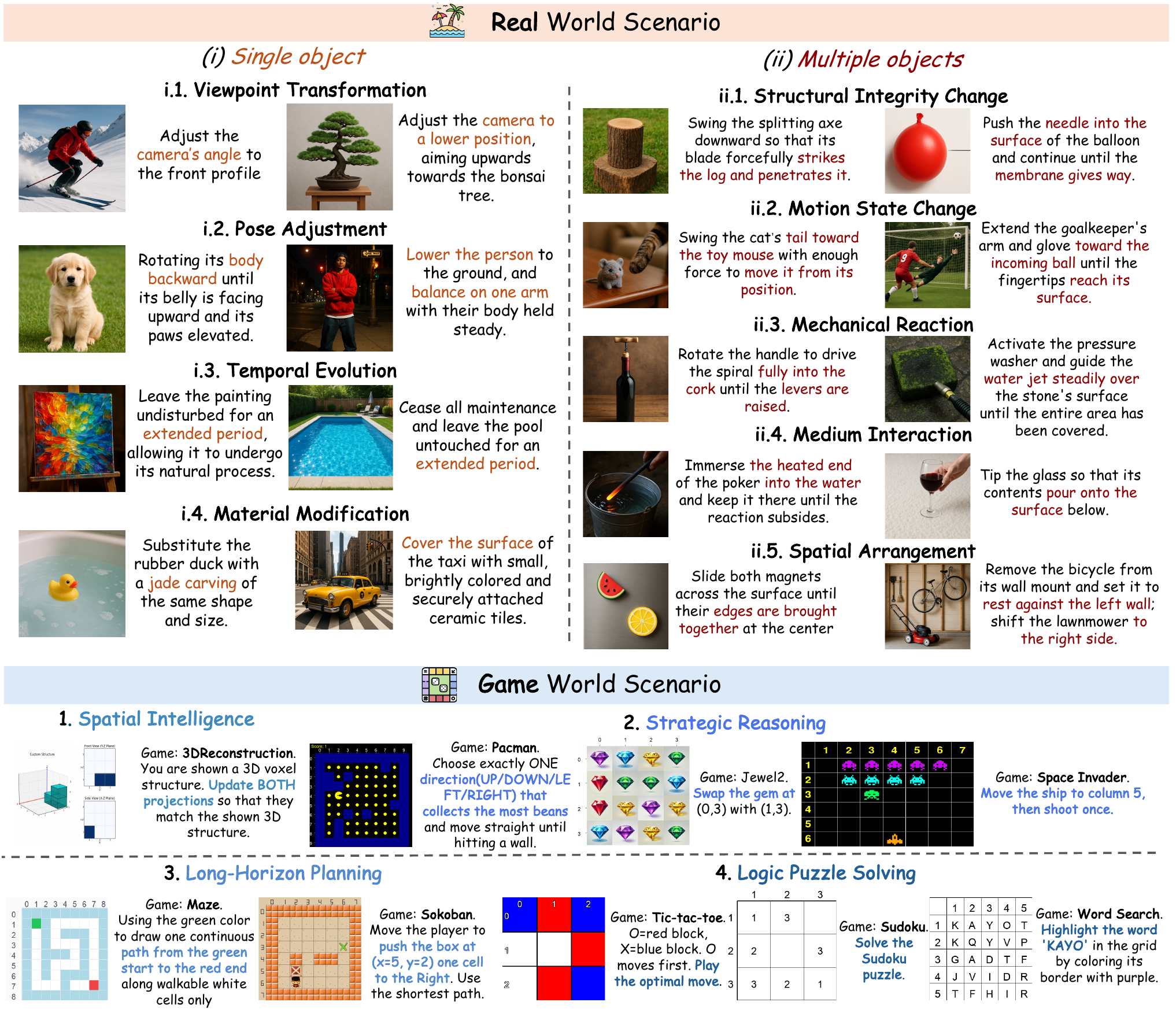} 
    \vspace{-0.2cm}
    \caption{Qualitative cases of evaluation dimensions in UniREditBench. We present qualitative examples for each dimension across both real-world and game-world scenarios.}
    \label{fig:unireditbench_categories}
    \vspace{-0.78cm}
\end{figure*}

\noindent\textbf{Reasoning-based Benchmarks for Image Generation and Editing.}
In T2I generation, several benchmarks \cite{wang2025unigenbench++,sun2025t2i,li2025easier,niu2025wise} have been developed to assess the reasoning capabilities of models in generating images. For example, WISE~\cite{niu2025wise} focuses on assessing models' world knowledge, such as cultural and physical understanding, while UniGenBench++~\cite{wang2025unigenbench++} unifies semantic generation evaluation, covering 10 primary dimensions and 27 sub-dimensions, such as logic reasoning, relational understanding, supporting multilingual and varying-length assessments.
In image editing evaluation, recent reasoning-based benchmarks~\cite{zhao2025envisioning,wu2025kris,jia2026compbench,wang2025complexbench} like RISEBench~\cite{zhao2025envisioning} aim to examine temporal, spatial, and logical editing capabilities of editing models. Besides, KRISBench~\cite{wu2025kris} introduces a knowledge-grounded taxonomy covering factual, conceptual, and procedural types. However, these benchmarks primarily focus on single-object knowledge and attribute reasoning. We posit that extending evaluation to multi-object interactions and scenarios governed by human-defined rules is a crucial next step. 
As for image quality evaluation \cite{xu2023imagereward,kirstain2023pick}, recent works like UnifiedReward~\cite{wang2025unified,wang2025unified-think} adopt the ``VLM-as-a-judge'' paradigm, leveraging the powerful capabilities of VLMs to score and provide explanatory judgments.
In image editing tasks, evaluation is more challenging because the evaluator needs to assess not only image quality but also understand complex editing instructions and final edited effects. Most studies like RISEBench and KRISBench utilize the property model~\cite{hurst2024gpt} to rate instruction following, temporal consistency, and image quality. Despite effectiveness, their evaluation relies mainly on textual references, which may lead to systematic misjudgments in complex reasoning tasks.

To this end, this work proposes UniREditBench, a unified reasoning-based image editing benchmark that spans a broad range of evaluation dimensions across real-world and game-world scenarios with multimodal dual-reference evaluation for more reliable and accurate assessments.

\section{UniREditBench}

\definecolor{UniRow}{RGB}{235,245,255}

\subsection{Overview}
With the rapid advancements in image editing models, existing benchmarks are gradually becoming less adequate to fully capture their comprehensive capabilities, particularly their reasoning-based editing abilities. Specifically, current benchmarks encounter two major challenges: \textbf{(1)} their evaluation primarily focuses on simple single-object attribute edits in real-world scenarios, neglecting complex multi-object interactions, as well as logical or strategic reasoning in game-world scenarios, where explicit human-defined rules govern the outcomes (Tab.\textcolor{red}{~\ref{tab:benchmark_compare}}); \textbf{(2)} their evaluation predominantly relies on CLIP-based metrics or VLM-based evaluators with text-only references, which may offer insufficient or inaccurate assessments, particularly in complex reasoning-intensive editing scenarios (Fig. \textcolor{red}{\ref{fig:eval_compare}}).

To this end, this work proposes \textbf{UniREditBench}, a unified reasoning-based image editing benchmark that covers a broad spectrum of reasoning dimensions in different scenarios. Compared with previous studies, this benchmark exhibits several key superiorities:

\begin{itemize}
    \item[\textbullet] \textbf{Broader scenario and reasoning dimension coverage.} It contains 2,700 high-quality samples organized into 8 primary reasoning dimensions and 18 sub-dimensions, spanning both real-world and game-world image editing tasks (Sec.\textcolor{red}{~\ref{sec:eva_dimen}}).
    \item[\textbullet] \textbf{Reliable dual-reference evaluation.} For each sample assessment, we introduce both the textual reference and ground-truth (GT) image reference. This multimodal reference enables VLM evaluators to perform direct and fine-grained comparisons at both the textual and visual levels with the generated images, leading to more reliable evaluation (Sec.\textcolor{red}{~\ref{sec:eval_pipe}}).
    \item[\textbullet] \textbf{Scalable multi-scenario data synthesis.} We propose an automatic data synthesis pipeline with distinct generation strategies tailored for real-world and game-world scenarios (Sec.\textcolor{red}{~\ref{sec:data_construct}}). 
\end{itemize}

\subsection{Evaluation Dimensions} \label{sec:eva_dimen}
In real-life applications, image editing scenarios often involve diverse requirements spanning both real and game-world contexts, where complex contextual understanding and implicit reasoning capabilities are crucial for accurate image edits. Therefore, UniREditBench organizes reasoning-based image editing tasks into a scenario-to-category hierarchy framework. As illustrated in Fig. \textcolor{red}{\ref{fig:teaser}}, it covers both real-world and game-world scenarios across 8 primary dimensions and 18 sub-categories, each representing a unique visual reasoning challenge with 150 human-inspected examples. We will elaborate on each dimension in the following.

\subsubsection{Real-World Scenarios}
Real-world scenarios involve editing tasks that reflect the perceptual and interaction dynamics observed in natural environments. These tasks involve transformations of individual objects or complex interactions among multiple objects. To handle such tasks, models must capture the semantic, physical, and temporal characteristics of objects, as well as their relationships.

\begin{enumerate}
    \item[\textbullet] \textbf{Single-Object Transformation} targets variations intrinsic to an individual object, including viewpoint and attribute changes that do not disrupt spatial relationships within the scene:
    \begin{itemize}
        \item[\textbullet] \textbf{Viewpoint Transformation:} Altering the perspective or viewing angle to exhibit alternative views of the same object (e.g., side, top-down, close-up).
        \item[\textbullet] \textbf{Pose Adjustment:} Modifying the articulation or positioning of an object's parts, such as limb configurations or postural shifts.
        \item[\textbullet] \textbf{Temporal Evolution:} Simulating natural progressions over time like aging, decay, or seasonal changes impacting the object's appearance.
        \item[\textbullet] \textbf{Material Modification:} Changing inherent surface or material properties (e.g., color, texture) while preserving geometry and location.
    \end{itemize}
    \item[\textbullet] \textbf{Multi-Object Interaction} involves mutual influences and state changes arising from the physical or spatial interactions among multiple objects:
    \begin{itemize}
        \item[\textbullet] \textbf{Structural Integrity Change:} Physical deformations resulting from forces or collisions.
        \item[\textbullet] \textbf{Motion State Change:} Dynamics induced by contact or force transmission leading to altered movement or posture.
        \item[\textbullet] \textbf{Mechanical Reaction:} State transitions caused by device operation or functional interactions.
        \item[\textbullet] \textbf{Medium Interaction:} Changes mediated by substances or environmental factors that affect appearance or state.
        \item[\textbullet] \textbf{Spatial Arrangement:} Reorganization or repositioning of multiple objects within the scene.
    \end{itemize}
\end{enumerate}

\subsubsection{Game-World Scenarios}
Game-world scenarios consist of tasks within synthetic environments governed by human-defined rules, evaluating logical, strategic, spatial, and long-horizon reasoning capabilities. These tasks require models to plan, deduce, and act in accordance with the explicit rules that govern the environment.
\begin{itemize}
    \item[\textbullet] \textbf{Long-Horizon Planning} requires multi-step sequential reasoning to accomplish distant goals, exemplified by navigation or puzzle games such as Maze-solving and Sokoban.
    \item[\textbullet] \textbf{Logical Puzzle Solving} involves constraint satisfaction and symbolic inference to produce valid solutions under formal rule sets, including Sudoku, Tic-tac-toe, and Word Search.
    \item[\textbullet] \textbf{Strategic Reasoning} requires resource management, adversarial planning over time, captured by games like Pacman, Jewel2, and Space Invader.
    \item[\textbullet] \textbf{Spatial Intelligence} focuses on geometric and topological reasoning within 3D environments, such as reconstructing spatial layouts in gaming contexts.
\end{itemize}
Representative examples are provided in Figs.\textcolor{red}{~\ref{fig:teaser}} and\textcolor{red}{~\ref{fig:unireditbench_categories}} to illustrate the scope and diversity of evaluation dimensions, and highlight the complexity and variety of tasks in our benchmark.

\subsection{Dual-Reference Evaluation}   \label{sec:eval_pipe} 
Evaluating reasoning-based image editing is intrinsically challenging due to the need for the evaluator to accurately understand the implicit reasoning intentions within the instruction. To achieve reliable and comprehensive assessments, we introduce a VLM-based multi-dimensional scoring schema, leveraging both textual and visual evaluation references. Specifically, for each sample, this pipeline evaluates three core dimensions:

\noindent \textbf{Instruction Following} measures how accurately the generated image reflects the input instruction, focusing on whether the explicit effect of the edit is properly manifested. Here, the VLM compares the output image $G$ against both the textual reference of edited effect $R_t$ and the corresponding reference GT image $R_i$ to verify compliance:
$$
S_{\mathrm{IF}} = \mathrm{VLM}(O, I, G, R_i, R_t)
$$
where $O$ represents the original image, $I$ denotes the editing instruction.

\noindent \textbf{Visual Consistency} assesses the preservation of image regions and attributes unrelated to the edit instruction, ensuring that changes are localized and do not inadvertently alter irrelevant scene elements. This criterion favors models capable of accurate, fine-grained editing rather than wholesale regeneration:
$$
S_{\mathrm{VC}} = \mathrm{VLM}(O, I, G)
$$

\noindent \textbf{Visual Quality} evaluates the realism of the generated output, checking for artifacts, distortions, and physical or logical implausibility in the final image:
$$
S_{\mathrm{VQ}} = \mathrm{VLM}(G)
$$
We choose GPT-4.1~\cite{hurst2024gpt} as the VLM evaluator. Each score $S$ ranges from 1 to 5, following prior detailed scoring guidelines~\cite{zhao2025envisioning,wu2025kris}. Finally, the overall evaluation score aggregates these via weighted sum:
$$
S_{\mathrm{Overall}} = \alpha_1 S_{\mathrm{IF}} + \alpha_2 S_{\mathrm{VC}} + \alpha_3 S_{\mathrm{VQ}},
$$
where $\alpha_1 = 0.5$, $\alpha_2 = 0.3$, and $\alpha_3 = 0.2$. 
This setting prioritizes instruction following to emphasize the importance of adhering to the instruction, and also incorporates visual consistency and quality, ensuring that areas unrelated to the instruction are preserved and that the overall image quality is maintained. Details of the weight selection strategy are provided in the Appendix.

\subsection{Multi-Scenario Data Synthesis} \label{sec:data_construct}
Given the distinct characteristics of real- and game-world contexts, we develop a specialized data generation process for each scenario, as illustrated in Fig.\textcolor{red}{~\ref{fig:data_construction}}. Detailed elaboration of each data synthesis process is provided below.

For \textbf{Real-World Scenario}, we employ a ``text-then-image'' data generation strategy. Specifically, \textbf{(1)} this process begins with hand-crafted textual triples that describe the original image, the editing instruction, and the textual reference of edited effect (a reasoning-based narrative of the anticipated outcome). Next, we use the powerful VLM~\cite{comanici2025gemini} to expand this initial set into a large corpus of text triples.
Subsequently, \textbf{(2)} these curated textual triples are input to GPT-4o~\cite{hurst2024gpt} to synthesize the original and edited images in alignment with the described textual reference of edited effect. \textbf{(3)} Finally, in the quality filtering stage, VLM~\cite{comanici2025gemini} is used to assess the generated images based on visual fidelity, instruction alignment, and potential hallucination risks. Additionally, it generates reasoning chain-of-thought (CoT) text for each qualified instance, ensuring the production of high-quality, reasoning-based image editing training data.

In \textbf{Game-World Scenario}, game states are inherently well-suited to be represented as structured reasoning-based editing data, where instructions can naturally be solved using Python code. Inspired by Game-RL~\cite{tong2025code2logic}, \textbf{(1)} we first design a diverse collection of game-based problems and develop corresponding Python programs tailored to each category. \textbf{(2)} Then, these programs automatically generate paired original and edited images, along with instructions, textual reference effects, and programmatic CoT reasoning traces. \textbf{(3)} To bridge the gap between programmatic and natural language CoT reasoning formats, we use VLMs to convert these reasoning traces into explanations that align with human inference patterns. Finally, quality filtering is applied to ensure the integrity and reliability of the data.

Overall, this multi-scenario data synthesis pipeline generates \textbf{UniREditBench}, a unified reasoning-based image editing benchmark, and \textbf{UniREdit-Data-100K}, a large-scale dataset with high-quality CoT annotations. We detail this dataset in the next section. 

\begin{figure}[t] 
    \centering
    \includegraphics[width=\textwidth]{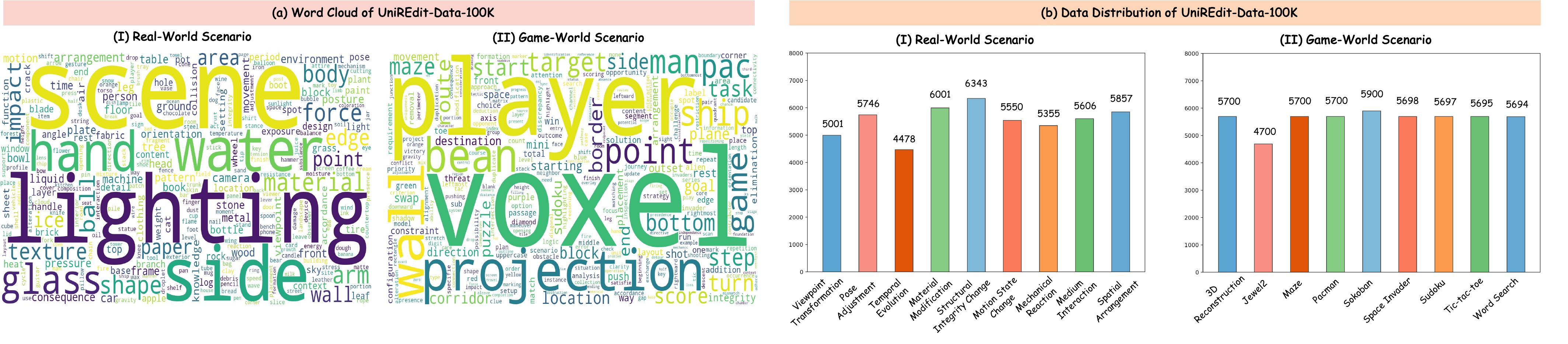} 
    \vspace{-0.7cm}
    \caption{Statistic visualization. We visualize (a) word clouds and (b) data distribution of our UniREdit-Data-100K.}
    \label{fig:word_cloud_data_distri}
    \vspace{-0.7cm}
\end{figure}

\section{UniREdit-Data-100K}
To enhance the capability of current generative models on reasoning-driven image editing, we propose UniREdit-Data-100K, which contains 100,421 samples spanning 8 reasoning dimensions and 18 categories defined in Sec. \textcolor{red}{\ref{sec:eva_dimen}}.

\subsection{Statistical Analysis}
UniREdit-Data-100K is designed with an emphasis on balance and diversity, ensuring that each reasoning category contains over 4,000 instances to effectively support model training across a wide range of editing tasks. It is divided into two primary scenarios: (i) Real-World Scenario, which captures natural object attributes and complex multi-object interactions, and (ii) Game-World Scenario, presenting structured, rule-based editing challenges, such as puzzles and strategic planning games.
We visualize the word cloud for both real-world and game-world subsets in Fig. \textcolor{red}{\ref{fig:word_cloud_data_distri}} (a) and the detailed distribution of samples across different categories in Fig. \textcolor{red}{\ref{fig:word_cloud_data_distri}} (b). These visualizations highlight the extensive vocabulary as well as the broad coverage across various categories of our dataset.

\begin{table*}[t]
\centering
\caption{In-domain quantitative comparisons on UniREditBench. \textit{GPT-4.1} is used as the evaluator. Best scores are in \textbf{bold}.}
\tiny
\label{tab:result_uni}
\vspace{-3mm}
\resizebox{\linewidth}{!}{%
\setlength{\tabcolsep}{2pt}
\begin{tabular}{l|ccccc|ccccc|c}
\toprule
\multirow{3}{*}{Model} &
\multicolumn{5}{c|}{\textbf{Real World Scenario}} &
\multicolumn{5}{c|}{\textbf{Game World Scenario}} \\
\cmidrule(lr){2-6}\cmidrule(lr){7-11}
 & Attribute & Structure & Physical  & Property & \multirow{2}{*}{Avg.}
 & Spatial & Strategic & Long-Horizon & Logic Puzzle & \multirow{2}{*}{Avg.} & \multirow{2}{*}{Overall} \\
 & Modification & Transform & Interaction & Response
 &
 & Intelligence & Reason & Plan & Solving
 &
 \\
\midrule

\multicolumn{12}{c}{\textbf{Closed-source Models}} 
\\
\midrule
FLUX-Kontext-Pro & 47.35 & 47.16 & 44.37 & 41.44 & 45.00 & 49.12 & 48.58 & 51.16 & 40.49 & 46.52 & 45.77 \\
Seedream4.0      & 69.54 & 73.13 & 67.88 & 62.40 & 68.20 & 39.27 & 43.54 & 43.79 & 51.91 & 45.91 & 57.03 \\
Wan2.5           & 74.66 & 69.74 & 63.51 & 62.87 & 67.23 & 63.73 & 47.13 & 55.00 & 54.93 & 52.67 & 61.36 \\
\raisebox{-0.2em}{\includegraphics[height=0.8em]{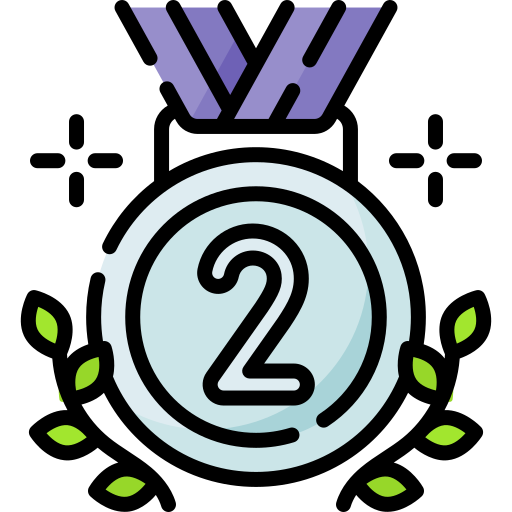}}Nano Banana      & 77.10 & 78.88 & 71.86 & 74.70 & 75.22 & 66.74 & \textbf{56.11} & 56.83 & \textbf{64.91} & 60.39 & 68.26 \\
\raisebox{-0.2em}{\includegraphics[height=0.8em]{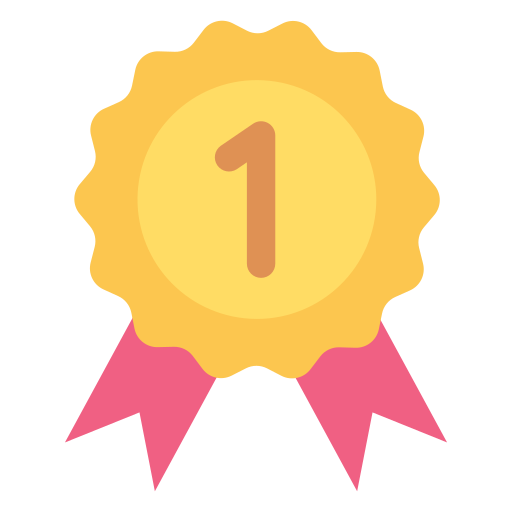}}GPT-4o           & \textbf{82.67} & \textbf{84.75} & \textbf{79.81} & \textbf{77.40} & \textbf{81.01} & \textbf{77.73} & 51.44 & \textbf{67.61} & 63.79 & \textbf{62.07} & \textbf{73.39} \\
\midrule 
\multicolumn{12}{c}{\textbf{Open-source Models}} \\
\midrule
MagicBrush       & 43.97 & 46.09 & 42.93 & 46.64 & 44.69 & 63.58 & 33.59 & 30.72 & 35.31 & 36.85 & 40.77 \\
Omnigen2         & 55.47 & 58.27 & 51.44 & 50.70 & 53.69 & 70.28 & 27.50 & 36.25 & 24.31 & 33.14 & 43.41 \\
Bagel            & 61.85 & 59.55 & 52.56 & 54.47 & 56.60 & 54.25 & 35.21 & 37.69 & 39.83 & 39.42 & 48.01 \\
Lumina-DiMOO     & 52.84 & 52.31 & 50.31 & 50.83 & 51.44 & 61.23 & 36.96 & 39.57 & 53.09 & 45.61 & 48.54 \\
Step1X-Edit      & 59.69 & 56.36 & 53.85 & 54.84 & 55.93 & 65.68 & 34.89 & 47.01 & 43.89 & 44.00 & 50.15 \\
Bagel-Think      & 61.45 & 59.51 & 52.65 & 55.68 & 56.80 & 66.29 & 43.23 & 43.00 & 41.30 & 45.10 & 50.96 \\
DreamOmni2       & 63.52 & 58.35 & 52.68 & 54.02 & 56.64 & 72.42 & 42.17 & 48.80 & 48.09 & 48.98 & 52.81 \\
UniWorld-V2      & 72.96 & 69.37 & 63.65 & 61.69 & 66.55 & 49.27 & 40.00 & 53.41 & 37.53 & 43.19 & 54.87 \\
\raisebox{-0.2em}{\includegraphics[height=0.8em]{images/second.png}}Qwen-Image-Edit  & 75.68 & 73.03 & 70.59 & 64.67 & 70.95 & 56.73 & 36.63 & 48.80 & 37.68 & 41.92 & 56.52 \\
\hline
\rowcolor[HTML]{E2F4E3}
\raisebox{-0.2em}{\includegraphics[height=0.8em]{images/winner.png}}\textbf{UniREdit-Bagel} (Ours)    & \textbf{76.73} & \textbf{77.80} & \textbf{76.57} & \textbf{71.44} & \textbf{75.74} & \textbf{84.90} & \textbf{72.83} & \textbf{84.88} & \textbf{83.72} & \textbf{80.48} & \textbf{78.15} \\
\bottomrule

\end{tabular}

}
\vspace{-0.4cm}
\end{table*}

\subsection{UniREdit-Bagel}
To further validate the effectiveness of our dataset, we use it to fine-tune Bagel \cite{deng2025bagel}, a unified understanding and generative model. Specifically, each training sample consists of the input image $O$, an editing instruction $I$, a stepwise CoT text $C$ that grounds the edit effects step by step, and the target edited image $G$. During training, the original image and instruction are first input into the model, which then generates a textual reasoning trace and synthesizes the edited image.
We supervise both the textual reasoning trace and the visual edit. Formally, for reasoning text supervision, we minimize the negative log-likelihood:
\[
\mathcal{L}_{\text{text}}
= - \sum^{T}_{t} \log p_\theta \big(y_t \mid y_{<t}, O, I \big).
\]
For image generation, we supervise the latent flow-matching loss~\cite{lipman2022flow} between the VAE latents of $O$ and $G$, conditioned on $(O, I, C)$:
\[
\mathcal{L}_{\text{img}} =
\mathbb{E}_{t \sim \mathcal{U}(0,1)}\;
\big\| u_\theta(z_t, t \,;\, O, I, C) - u^\star(z_t, t) \big\|_2^2,
\]
where $u_\theta$ is the learned time-conditioned velocity field on the latent path from $z_O$ to $z_G$, and $u^\star$ is the target velocity. Finally, the overall objective is
\[
\mathcal{L} = \lambda_{\text{text}} \mathcal{L}_{\text{text}} + \lambda_{\text{img}} \mathcal{L}_{\text{img}}.
\]
\noindent 
Under the influence of \(\mathcal{L}_{\text{text}}\), the model enhances its reasoning ability through explicit CoT learning, which effectively guides the accurate image editing, while 
\(\mathcal{L}_{\text{img}}\) improves both the correctness and fidelity of the edited image.

\section{Experiment}

\definecolor{UniRow}{RGB}{235,245,255} 

\begin{figure*}[t]
    \centering
    \includegraphics[width=0.85\textwidth]{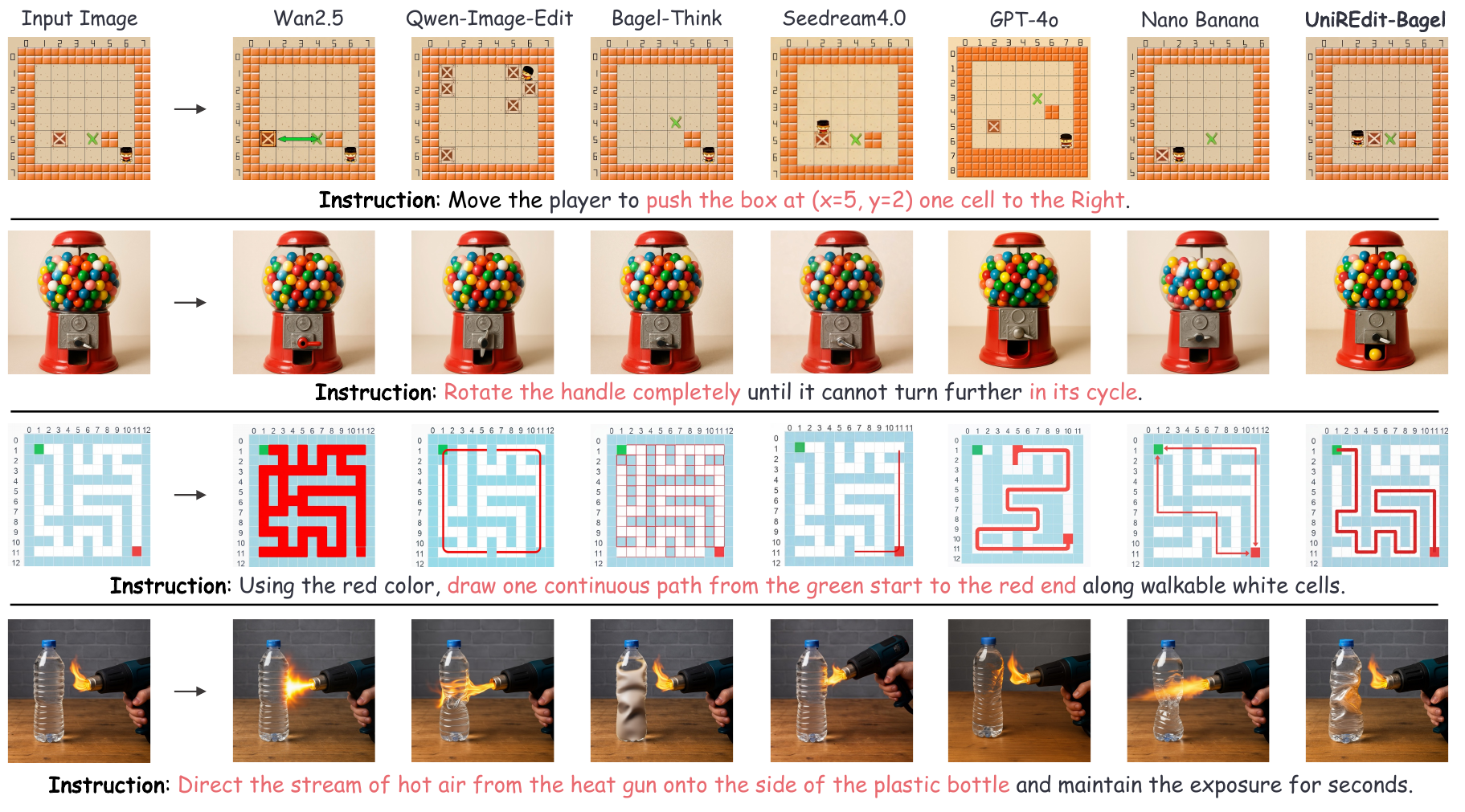} 
    \caption{Qualitative editing result comparison. Our UniREdit-Bagel demonstrates significant superiority in both instruction following and visual quality compared with state-of-the-art closed-source and open-source models.}
    \label{fig:uniredite_qualitative}
    \vspace{-0.5cm}
\end{figure*}

\subsection{Implementation Details}
\textbf{Baselines.} We benchmark \textit{closed-source} models including: GPT-4o~\cite{hurst2024gpt}, Nano Banana~\cite{Gemini-2.5-Flash-Image}, Gemini-2.0~\cite{comanici2025gemini}, Seedream4.0~\cite{seedream4}, Wan2.5~\cite{wan2025wan}, and FLUX-Kontext-Pro~\cite{flux_kontext}, as well as \textit{open-source} models including: Bagel~\cite{deng2025bagel}, Qwen-Image-Edit~\cite{qwen-image}, Step1X-Edit~\cite{liu2025step1x}, FLUX.1-Kontext-dev~\cite{flux_kontext}, Emu2~\cite{sun2024generative}, Omnigen2~\cite{wu2025omnigen2}, Omnigen~\cite{xiao2025omnigen}, HiDream-Edit~\cite{cai2025hidream}, MagicBrush~\cite{zhang2023magicbrush}, and AnyEdit~\cite{yu2025anyedit}.

\noindent\textbf{Training and Evaluation.}
We train all Bagel~\cite{deng2025bagel} components, except the VAE, for 5,000 iterations on UniREdit-Data-100K using the Adam optimizer and a cosine learning-rate schedule with 500 warm-up steps, a peak learning rate of $2\times10^{-5}$, and a minimum learning rate of $10^{-6}$. The loss weights are set to $\lambda_{\text{text}}=2$ and $\lambda_{\text{img}}=1$. During inference, we use the official inference settings provided by Bagel. To ensure fair comparisons with other baselines, we adopt the original inference configurations of these models.

\subsection{Benchmarking Results on UniREditBench}

As shown in Tab.\textcolor{red}{~\ref{tab:result_uni}}, among \textbf{closed-source} models, GPT-4o achieves the highest overall performance across all scenarios, with Nano Banana performing comparably. Wan2.5 delivers balanced results on real-world tasks but lags on game scenarios that require strategic reasoning. Besides, Seedream4.0 is competitive on \textit{structure transform} yet encounters challenges in game scenarios. Among \textbf{open-source} baselines, Qwen-Image-Edit performs strongly on real-world tasks such as \textit{attribute modification} and \textit{structure transform}. However, most models remain comparatively weak on game scenarios like \textit{Strategic Reasoning}. \textbf{Overall}, compared with open-source methods, closed-source models, particularly GPT-4o, maintain a clear advantage. While some open-source models are competitive on specific real-world tasks, they generally struggle with complex reasoning in game scenarios. Notably, only GPT-4o and Nano Banana achieve an average score greater than 60 on game scenarios, underscoring that this setting remains highly challenging and serves as a useful test for current models.

\subsection{Comparison Results of UniREdit-Bagel}

\begin{table}[t!]
\centering
\caption{Out-of-distribution quantitative performance comparison on RISEBench~\cite{zhao2025envisioning}. \textit{GPT-4.1} is used as the evaluator. Best scores are in \textbf{bold}.}
\vspace{-0.2cm}
\label{tab:result_rise}
\vspace{0.3em}
\scriptsize
\resizebox{0.8\linewidth}{!}{%
\begin{tabular}{l|cccc|c}
\toprule
Models \rule{0pt}{0.25em} & Temporal & Causal & Spatial & Logical & \textbf{Overall} \rule[-0.3em]{0pt}{1.5em}\\
\midrule

\multicolumn{6}{c}{\textbf{Closed-source Models}} \\
\midrule
Gemini-2.0-Flash-pre & 10.6\%          & 13.3\%          & 11\%            & 2.3\%           & 9.4\% \\
Seedream4.0 & 12.9\%	& 12.2\%	& 11.0\%	& 7.1\%	& 10.8\% \\
Gemini-2.0-Flash-exp & 8.2\%           & 15.5\%          & 23.0\%          & 4.7\%           & 13.3\% \\
\raisebox{-0.2em}{\includegraphics[height=0.9em]{images/second.png}}GPT-4o         & \textbf{34.1\%} & 32.2\% & \textbf{37.0\%} & 10.6\% & 28.9\% \\
\raisebox{-0.2em}{\includegraphics[height=0.9em]{images/winner.png}}Nano Banana & 25.9\% & \textbf{47.8\%} & \textbf{37.0\%} & \textbf{18.8\%} & \textbf{32.8\%} \\
\midrule

\multicolumn{6}{c}{\textbf{Open-source Models}} \\
\midrule
HiDream-Edit     & 0.0\% & 0.0\% & 0.0\% & 0.0\% & 0.0\% \\
OmniGen     & 1.2\% & 1.0\% & 0.0\% & 1.2\% & 0.8\% \\
Step1X-Edit & 0.0\% & 2.2\% & 2\%   & 3.5\% & 1.9\% \\
Bagel       & 3.5\% & 4.4\% & 9.0\% & 5.9\% & 5.8\% \\
FLUX.1-Kontext-Dev	& 2.3\%	& 5.5\%	& 13.0\%	& 1.2\%	& 5.8\% \\
Qwen-Image-Edit	& 4.7\%	& 10.0\%	& 17.0\%	& 2.4\%	& 8.9\% \\
\raisebox{-0.2em}{\includegraphics[height=0.9em]{images/second.png}}Bagel-Think & 4.7\% & 15.5\% & 14.0\% & 1.2\% & 9.2\% \\
\hline
\rowcolor[HTML]{E2F4E3}
\raisebox{-0.2em}{\includegraphics[height=0.9em]{images/winner.png}}\textbf{UniREdit-Bagel} (Ours) & \textbf{22.4\%} & \textbf{18.9\%} & \textbf{21.0\%} & \textbf{10.6\%} & \textbf{18.3\%} \\
\bottomrule
\end{tabular}
}
\vspace{-0.4cm}
\end{table}

\noindent\textbf{Quantitative.}
UniREdit-Bagel achieves the best overall performance among all closed- and open-source models on UniREditBench, surpassing the second-place GPT-4o by a substantial margin. The largest gains occur in game-world scenarios (+17.08), indicating exceptional capability of understanding and processing complex reasoning image editing tasks.
In out-of-distribution performance comparison, 
UniREdit-Bagel achieves the strongest open-source results across all categories on RISEBench, shown in Tab. \textcolor{red}{\ref{tab:result_rise}}, improving upon the Bagel-Think baseline by 9.1 points and surpassing the closed-source Gemini-2.0-Flash-exp by 5.0 points. It also remains competitive with top closed-source models like Nano Banana, narrowing the gap between open- and closed-source models.

\noindent\textbf{Qualitative.}
The qualitative results presented in Fig.\textcolor{red}{~\ref{fig:uniredite_qualitative}} highlight the strengths of our UniREdit-Bagel across various tasks. Specifically, in Row 4, most models fail to reliably reproduce the physical effect induced by the heat. Although several baselines (e.g., Nano Banana and Qwen-Image-Edit) capture the heat-induced warping of a plastic bottle under sustained heat gun exposure, they fail to preserve the heat trace. Notably, UniREdit-Bagel not only renders the deformation accurately but also preserves the heat trace, offering superior visual consistency.
Besides, in the Sokoban and Maze game settings (Rows 1 and 3), most baselines (e.g. Seedream4.0, Nano Banana, and Wan2.5) struggle with box pushing and maze completion. In contrast, UniREdit-Bagel excels in both fulfilling the instruction and maintaining the coherence of unrelated content.

\subsection{Discussion and Analysis}

\subsubsection{VLM-as-Judge Assessment}
To assess the reliability of using VLMs as evaluators, we conducted a user study with six human experts on 200 randomly selected samples across four models (Qwen-Image-Edit, Nano Banana, GPT-4o, UniREdit-Bagel). We reported the Mean Absolute Error (MAE) between experts' scores and GPT-4.1 scores in Tab.\textcolor{red}{~\ref{tab:mae_right}}. It is demonstrated that the average MAE of each score and model falls below 1, indicating that scores by GPT-4.1 are closely aligned with those predicted by human experts.

\vspace{-0.2cm}

\subsubsection{Ablation on Dual-Reference Evaluation Protocol}
\begin{figure*}[t]
\centering
\resizebox{0.9\linewidth}{!}{%
\begin{minipage}[t]{0.53\textwidth}
\vspace{0pt}
\centering
\captionof{figure}{Mean Absolute Error (MAE) between human and VLM scores under different reference settings.}
\vspace{-5pt}
\label{fig:mae_left}
\includegraphics[width=\linewidth]{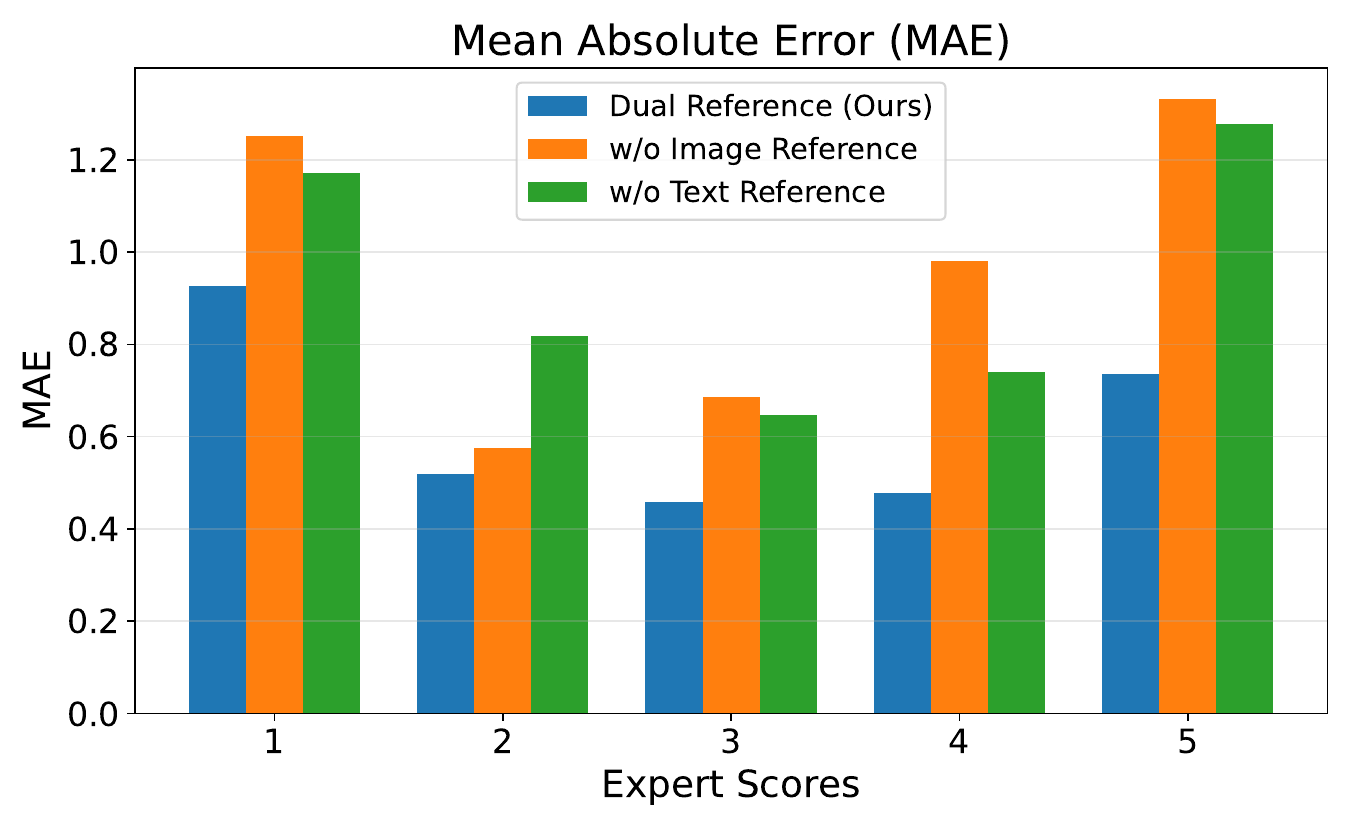}
\end{minipage}
\hfill
\begin{minipage}[t]{0.45\textwidth}
\vspace{-11pt}
\centering
\captionof{table}{MAE between human and VLM scores across Instruction Following (IF), Visual Consistency (VC), and Visual Quality (VQ).}
\vspace{7pt}
\label{tab:mae_right}
\resizebox{\linewidth}{!}{%
\begin{tabular}{lcccc}
\toprule
Model & IF & VC & VQ & Avg. \\
\midrule
Qwen-Image-Edit & 0.73 & 0.62 & 0.47 & 0.61 \\
Nano Banana     & 0.85 & 0.42 & 0.66 & 0.64 \\
GPT-4o          & 0.42 & 0.68 & 0.22 & 0.44 \\
UniREdit-Bagel  & 0.66 & 0.54 & 0.51 & 0.57 \\
\midrule
Avg.            & 0.67 & 0.57 & 0.47 &      \\
\bottomrule
\end{tabular}}
\end{minipage}
}

\vspace{-0.5cm}
\end{figure*}

We conduct an ablation study under different reference settings. Compared with settings that omit the image or the text reference (Fig.\textcolor{red}{~\ref{fig:mae_left}}), our dual-reference protocol consistently achieves the lowest MAE between human and VLM scores across all expert score levels, validating the accuracy of our evaluation pipeline.

\vspace{-0.7cm}

\par\vspace{\baselineskip}
\subsubsection{More Analysis} We provide additional analyses in the Appendix, including \textbf{(1)} an ablation study of the score weighting parameters, showing that $(\alpha_1,\alpha_2,\alpha_3)=(0.50,0.30,0.20)$ produces scores most consistent with human experts and is adopted as the weighting scheme;
\textbf{(2)} quantitative results on KRISBench, showing that training on UniREdit-Data-100K improves out-of-distribution generalization on other benchmarks;
\textbf{(3)} consistent scoring trends between Qwen3VL-32B-Instruct and GPT-4.1;
\textbf{(4)} an ablation study on game-world data, demonstrating transferability to real-world scenarios;
\textbf{(5)} investigations into the role of chain-of-thought (CoT) reasoning;
\textbf{(6)} data-scaling analysis, confirming consistent gains from 20K to 100K samples; and
\textbf{(7)} additional qualitative comparisons and model-specific failure modes.

\section{Conclusion}
This paper presents \textbf{UniREditBench}, a unified reasoning-based benchmark for image editing with broad dimension coverage and a robust dual-reference evaluation protocol. We further introduce a multi-scenario synthesis pipeline and release \textbf{UniREdit-Data-100K}, a large-scale dataset with high-quality CoT annotations. To demonstrate its effectiveness, we fine-tune Bagel on this dataset and obtain \textbf{UniREdit-Bagel}, which achieves substantial quantitative and qualitative gains. Comprehensive benchmarking of open-source and closed-source image editing models reveals their strengths and weaknesses across various aspects.

\section*{Acknowledgments}
This work is supported by Shanghai Innovation Institute.

{
    \small
    \bibliographystyle{splncs04}
    \bibliography{main}

@String(CVPR= {IEEE Conf. Comput. Vis. Pattern Recog.})

@String(AAAI = {AAAI})

@String(CVPR  = {CVPR})

@article{deng2025bagel,
  title   = {Emerging Properties in Unified Multimodal Pretraining},
  author  = {Deng, Chaorui and Zhu, Deyao and Li, Kunchang and Gou, Chenhui and Li, Feng and Wang, Zeyu and Zhong, Shu and Yu, Weihao and Nie, Xiaonan and Song, Ziang and Shi, Guang and Fan, Haoqi},
  journal = {arXiv preprint arXiv:2505.14683},
  year    = {2025}
}

@article{hurst2024gpt,
  title={Gpt-4o system card},
  author={Hurst, Aaron and Lerer, Adam and Goucher, Adam P and Perelman, Adam and Ramesh, Aditya and Clark, Aidan and Ostrow, AJ and Welihinda, Akila and Hayes, Alan and Radford, Alec and others},
  journal={arXiv preprint arXiv:2410.21276},
  year={2024}
}

@article{wu2025kris,
  title={KRIS-Bench: Benchmarking Next-Level Intelligent Image Editing Models},
  author={Wu, Yongliang and Li, Zonghui and Hu, Xinting and Ye, Xinyu and Zeng, Xianfang and Yu, Gang and Zhu, Wenbo and Schiele, Bernt and Yang, Ming-Hsuan and Yang, Xu},
  journal={arXiv preprint arXiv:2505.16707},
  year={2025}
}

@article{liu2025step1x,
  title={Step1x-edit: A practical framework for general image editing},
  author={Liu, Shiyu and Han, Yucheng and Xing, Peng and Yin, Fukun and Wang, Rui and Cheng, Wei and Liao, Jiaqi and Wang, Yingming and Fu, Honghao and Han, Chunrui and others},
  journal={arXiv preprint arXiv:2504.17761},
  year={2025}
}

@inproceedings{yu2025anyedit,
  title={Anyedit: Mastering unified high-quality image editing for any idea},
  author={Yu, Qifan and Chow, Wei and Yue, Zhongqi and Pan, Kaihang and Wu, Yang and Wan, Xiaoyang and Li, Juncheng and Tang, Siliang and Zhang, Hanwang and Zhuang, Yueting},
  booktitle={CVPR},
  pages={26125--26135},
  year={2025}
}

@inproceedings{xiao2025omnigen,
  title={Omnigen: Unified image generation},
  author={Xiao, Shitao and Wang, Yueze and Zhou, Junjie and Yuan, Huaying and Xing, Xingrun and Yan, Ruiran and Li, Chaofan and Wang, Shuting and Huang, Tiejun and Liu, Zheng},
  booktitle={CVPR},
  pages={13294--13304},
  year={2025}
}

@article{shen2024imagpose,
  title={Imagpose: A unified conditional framework for pose-guided person generation},
  author={Shen, Fei and Tang, Jinhui},
  journal={Advances in neural information processing systems},
  volume={37},
  pages={6246--6266},
  year={2024}
}

@article{cai2025hidream,
  title={HiDream-I1: A High-Efficient Image Generative Foundation Model with Sparse Diffusion Transformer},
  author={Cai, Qi and Chen, Jingwen and Chen, Yang and Li, Yehao and Long, Fuchen and Pan, Yingwei and Qiu, Zhaofan and Zhang, Yiheng and Gao, Fengbin and Xu, Peihan and others},
  journal={arXiv preprint arXiv:2505.22705},
  year={2025}
}

@misc{Gemini-2.5-Flash-Image,
  title={Introducing Gemini 2.5 Flash Image, our state-of-the-art image model},
  author={Google},
  howpublished={\url{https://developers.googleblog.com/en/introducing-gemini-2-5-flash-image/}},
  year={2025}
}

@misc{flux_kontext,
      title={FLUX.1 Kontext: Flow Matching for In-Context Image Generation and Editing in Latent Space},
      author={Batifol, Stephen and Blattmann, Andreas and Boesel, Frederic and Consul, Saksham and Diagne, Cyril and Dockhorn, Tim and English, Jack and English, Zion and Esser, Patrick and Kulal, Sumith and Lacey, Kyle and Levi, Yam and Li, Cheng and Lorenz, Dominik and M\"uller, Jonas and Podell, Dustin and Rombach, Robin and Saini, Harry and Sauer, Axel and Smith, Luke},
      year={2025},
      eprint={2506.15742},
      archivePrefix={arXiv},
      primaryClass={cs.GR},
      url={https://arxiv.org/abs/2506.15742},
}

@article{qwen-image,
  title={Qwen-image technical report},
  author={Wu, Chenfei and Li, Jiahao and Zhou, Jingren and Lin, Junyang and Gao, Kaiyuan and Yan, Kun and Yin, Sheng-ming and Bai, Shuai and Xu, Xiao and Chen, Yilei and others},
  journal={arXiv preprint arXiv:2508.02324},
  year={2025}
}

@article{seedream4,
  title={Seedream 4.0},
  author={ByteDance Seed},
  journal={\url{https://seed.bytedance.com/en/seedream4_0}},
  year={2025}
}

@article{zhao2025envisioning,
  title={Envisioning beyond the pixels: Benchmarking reasoning-informed visual editing},
  author={Zhao, Xiangyu and Zhang, Peiyuan and Tang, Kexian and Zhu, Xiaorong and Li, Hao and Chai, Wenhao and Zhang, Zicheng and Xia, Renqiu and Zhai, Guangtao and Yan, Junchi and others},
  journal={arXiv preprint arXiv:2504.02826},
  year={2025}
}

@inproceedings{huang2024smartedit,
  title={Smartedit: Exploring complex instruction-based image editing with multimodal large language models},
  author={Huang, Yuzhou and Xie, Liangbin and Wang, Xintao and Yuan, Ziyang and Cun, Xiaodong and Ge, Yixiao and Zhou, Jiantao and Dong, Chao and Huang, Rui and Zhang, Ruimao and others},
  booktitle={CVPR},
  pages={8362--8371},
  year={2024}
}

@article{wu2025omnigen2,
  title={OmniGen2: Exploration to Advanced Multimodal Generation},
  author={Wu, Chenyuan and Zheng, Pengfei and Yan, Ruiran and Xiao, Shitao and Luo, Xin and Wang, Yueze and Li, Wanli and Jiang, Xiyan and Liu, Yexin and Zhou, Junjie and others},
  journal={arXiv preprint arXiv:2506.18871},
  year={2025}
}

@inproceedings{shen2025imagdressing,
  title={Imagdressing-v1: Customizable virtual dressing},
  author={Shen, Fei and Jiang, Xin and He, Xin and Ye, Hu and Wang, Cong and Du, Xiaoyu and Li, Zechao and Tang, Jinhui},
  booktitle={Proceedings of the AAAI Conference on Artificial Intelligence},
  volume={39},
  number={7},
  pages={6795--6804},
  year={2025}
}

@article{zhang2023magicbrush,
  title={Magicbrush: A manually annotated dataset for instruction-guided image editing},
  author={Zhang, Kai and Mo, Lingbo and Chen, Wenhu and Sun, Huan and Su, Yu},
  journal={NeurIPS},
  volume={36},
  pages={31428--31449},
  year={2023}
}

@inproceedings{brooks2023instructpix2pix,
  title={Instructpix2pix: Learning to follow image editing instructions},
  author={Brooks, Tim and Holynski, Aleksander and Efros, Alexei A},
  booktitle={CVPR},
  pages={18392--18402},
  year={2023}
}

@article{li2025uniworld,
  title={Uniworld-V2: Reinforce Image Editing with Diffusion Negative-aware Finetuning and MLLM Implicit Feedback},
  author={Li, Zongjian and Liu, Zheyuan and Zhang, Qihui and Lin, Bin and Yuan, Shenghai and Yan, Zhiyuan and Ye, Yang and Yu, Wangbo and Niu, Yuwei and Yuan, Li},
  journal={arXiv preprint arXiv:2510.16888},
  year={2025}
}

@article{xiao2025mindomni,
  title={Mindomni: Unleashing reasoning generation in vision language models with rgpo},
  author={Xiao, Yicheng and Song, Lin and Chen, Yukang and Luo, Yingmin and Chen, Yuxin and Gan, Yukang and Huang, Wei and Li, Xiu and Qi, Xiaojuan and Shan, Ying},
  journal={arXiv preprint arXiv:2505.13031},
  year={2025}
}

@article{ye2025imgedit,
  title={Imgedit: A unified image editing dataset and benchmark},
  author={Ye, Yang and He, Xianyi and Li, Zongjian and Lin, Bin and Yuan, Shenghai and Yan, Zhiyuan and Hou, Bohan and Yuan, Li},
  journal={arXiv preprint arXiv:2505.20275},
  year={2025}
}

@article{zhang2025r,
  title={R-Genie: Reasoning-Guided Generative Image Editing},
  author={Zhang, Dong and He, Lingfeng and Yan, Rui and Shen, Fei and Tang, Jinhui},
  journal={arXiv preprint arXiv:2505.17768},
  year={2025}
}

@article{he2025reasoning,
  title={Reasoning to Edit: Hypothetical Instruction-Based Image Editing with Visual Reasoning},
  author={He, Qingdong and Chen, Xueqin and Wang, Chaoyi and Pan, Yanjie and Hu, Xiaobin and Gan, Zhenye and Wang, Yabiao and Wang, Chengjie and Li, Xiangtai and Zhang, Jiangning},
  journal={arXiv preprint arXiv:2507.01908},
  year={2025}
}

@article{tong2025code2logic,
  title={Code2Logic: Game-Code-Driven Data Synthesis for Enhancing VLMs General Reasoning},
  author={Tong, Jingqi and Tang, Jixin and Li, Hangcheng and Mou, Yurong and Zhang, Ming and Zhao, Jun and Wen, Yanbo and Song, Fan and Zhan, Jiahao and Lu, Yuyang and others},
  journal={arXiv preprint arXiv:2505.13886},
  year={2025}
}

@article{li2024vcbench,
  title={Vcbench: A controllable benchmark for symbolic and abstract challenges in video cognition},
  author={Li, Chenglin and Chen, Qianglong and Li, Zhi and Tao, Feng and Zhang, Yin},
  journal={arXiv preprint arXiv:2411.09105},
  year={2024}
}

@article{comanici2025gemini,
  title={Gemini 2.5: Pushing the frontier with advanced reasoning, multimodality, long context, and next generation agentic capabilities},
  author={Comanici, Gheorghe and Bieber, Eric and Schaekermann, Mike and Pasupat, Ice and Sachdeva, Noveen and Dhillon, Inderjit and Blistein, Marcel and Ram, Ori and Zhang, Dan and Rosen, Evan and others},
  journal={arXiv preprint arXiv:2507.06261},
  year={2025}
}

@inproceedings{wang2025dreamtext,
  title={DreamText: High Fidelity Scene Text Synthesis},
  author={Wang, Yibin and Zhang, Weizhong and Xu, Honghui and Jin, Cheng},
  booktitle={CVPR},
  pages={28555--28563},
  year={2025}
}

@article{wang2025pref,
  title={Pref-grpo: Pairwise preference reward-based grpo for stable text-to-image reinforcement learning},
  author={Wang, Yibin and Li, Zhimin and Zang, Yuhang and Zhou, Yujie and Bu, Jiazi and Wang, Chunyu and Lu, Qinglin and Jin, Cheng and Wang, Jiaqi},
  journal={arXiv preprint arXiv:2508.20751},
  year={2025}
}

@inproceedings{gong2026freeinpaint,
  title={FreeInpaint: Tuning-free Prompt Alignment and Visual Rationality Enhancement in Image Inpainting},
  author={Gong, Chao and Li, Dong and Pan, Yingwei and Chen, Jingjing and Yao, Ting and Mei, Tao},
  booktitle={Proceedings of the AAAI Conference on Artificial Intelligence},
  volume={40},
  number={6},
  pages={4239--4247},
  year={2026}
}

@article{niu2025wise,
  title={Wise: A world knowledge-informed semantic evaluation for text-to-image generation},
  author={Niu, Yuwei and Ning, Munan and Zheng, Mengren and Jin, Weiyang and Lin, Bin and Jin, Peng and Liao, Jiaqi and Feng, Chaoran and Ning, Kunpeng and Zhu, Bin and others},
  journal={arXiv preprint arXiv:2503.07265},
  year={2025}
}

@inproceedings{jia2026compbench,
  title={Compbench: Benchmarking complex instruction-guided image editing},
  author={Jia, Bohan and Huang, Wenxuan and Tang, Yuntian and Qiao, Junbo and Liao, Jincheng and Cao, Shaosheng and Zhao, Fei and Feng, Zhaopeng and Gu, Zhouhong and Yin, Zhenfei and others},
  booktitle={Proceedings of the IEEE/CVF Conference on Computer Vision and Pattern Recognition},
  pages={1112--1122},
  year={2026}
}

@inproceedings{wang2024primecomposer,
  title={Primecomposer: Faster progressively combined diffusion for image composition with attention steering},
  author={Wang, Yibin and Zhang, Weizhong and Zheng, Jianwei and Jin, Cheng},
  booktitle={ACM MM},
  pages={10824--10832},
  year={2024}
}

@article{xin2025lumina,
  title={Lumina-dimoo: An omni diffusion large language model for multi-modal generation and understanding},
  author={Xin, Yi and Qin, Qi and Luo, Siqi and Zhu, Kaiwen and Yan, Juncheng and Tai, Yan and Lei, Jiayi and Cao, Yuewen and Wang, Keqi and Wang, Yibin and others},
  journal={arXiv preprint arXiv:2510.06308},
  year={2025}
}

@article{lipman2022flow,
  title={Flow matching for generative modeling},
  author={Lipman, Yaron and Chen, Ricky TQ and Ben-Hamu, Heli and Nickel, Maximilian and Le, Matt},
  journal={arXiv preprint arXiv:2210.02747},
  year={2022}
}

@article{wan2025wan,
  title={Wan: Open and advanced large-scale video generative models},
  author={Wan, Team and Wang, Ang and Ai, Baole and Wen, Bin and Mao, Chaojie and Xie, Chen-Wei and Chen, Di and Yu, Feiwu and Zhao, Haiming and Yang, Jianxiao and others},
  journal={arXiv preprint arXiv:2503.20314},
  year={2025}
}

@inproceedings{sun2024generative,
  title={Generative multimodal models are in-context learners},
  author={Sun, Quan and Cui, Yufeng and Zhang, Xiaosong and Zhang, Fan and Yu, Qiying and Wang, Yueze and Rao, Yongming and Liu, Jingjing and Huang, Tiejun and Wang, Xinlong},
  booktitle={CVPR},
  pages={14398--14409},
  year={2024}
}

@article{wang2024lift,
  title={Lift: Leveraging human feedback for text-to-video model alignment},
  author={Wang, Yibin and Tan, Zhiyu and Wang, Junyan and Yang, Xiaomeng and Jin, Cheng and Li, Hao},
  journal={arXiv preprint arXiv:2412.04814},
  year={2024}
}

@inproceedings{kawar2023imagic,
  title={Imagic: Text-based real image editing with diffusion models},
  author={Kawar, Bahjat and Zada, Shiran and Lang, Oran and Tov, Omer and Chang, Huiwen and Dekel, Tali and Mosseri, Inbar and Irani, Michal},
  booktitle={CVPR},
  pages={6007--6017},
  year={2023}
}

@article{meng2021sdedit,
  title={Sdedit: Guided image synthesis and editing with stochastic differential equations},
  author={Meng, Chenlin and He, Yutong and Song, Yang and Song, Jiaming and Wu, Jiajun and Zhu, Jun-Yan and Ermon, Stefano},
  journal={arXiv preprint arXiv:2108.01073},
  year={2021}
}

@article{hertz2022prompt,
  title={Prompt-to-prompt image editing with cross attention control},
  author={Hertz, Amir and Mokady, Ron and Tenenbaum, Jay and Aberman, Kfir and Pritch, Yael and Cohen-Or, Daniel},
  journal={arXiv preprint arXiv:2208.01626},
  year={2022}
}

@inproceedings{avrahami2022blended,
  title={Blended diffusion for text-driven editing of natural images},
  author={Avrahami, Omri and Lischinski, Dani and Fried, Ohad},
  booktitle={CVPR},
  pages={18208--18218},
  year={2022}
}

@inproceedings{kim2022diffusionclip,
  title={Diffusionclip: Text-guided diffusion models for robust image manipulation},
  author={Kim, Gwanghyun and Kwon, Taesung and Ye, Jong Chul},
  booktitle={CVPR},
  pages={2426--2435},
  year={2022}
}

@inproceedings{zhang2024hive,
  title={Hive: Harnessing human feedback for instructional visual editing},
  author={Zhang, Shu and Yang, Xinyi and Feng, Yihao and Qin, Can and Chen, Chia-Chih and Yu, Ning and Chen, Zeyuan and Wang, Huan and Savarese, Silvio and Ermon, Stefano and others},
  booktitle={CVPR},
  pages={9026--9036},
  year={2024}
}

@article{he2024freeedit,
  title={Freeedit: Mask-free reference-based image editing with multi-modal instruction},
  author={He, Runze and Ma, Kai and Huang, Linjiang and Huang, Shaofei and Gao, Jialin and Wei, Xiaoming and Dai, Jiao and Han, Jizhong and Liu, Si},
  journal={arXiv preprint arXiv:2409.18071},
  year={2024}
}

@article{wang2024emu3,
  title={Emu3: Next-token prediction is all you need},
  author={Wang, Xinlong and Zhang, Xiaosong and Luo, Zhengxiong and Sun, Quan and Cui, Yufeng and Wang, Jinsheng and Zhang, Fan and Wang, Yueze and Li, Zhen and Yu, Qiying and others},
  journal={arXiv preprint arXiv:2409.18869},
  year={2024}
}

@article{han2025controlthinker,
  title={ControlThinker: Unveiling Latent Semantics for Controllable Image Generation through Visual Reasoning},
  author={Han, Feng and Jiao, Yang and Chen, Shaoxiang and Xu, Junhao and Chen, Jingjing and Jiang, Yu-Gang},
  journal={arXiv preprint arXiv:2506.03596},
  year={2025}
}

@article{wang2025unigenbench++,
  title={UniGenBench++: A Unified Semantic Evaluation Benchmark for Text-to-Image Generation},
  author={Wang, Yibin and Li, Zhimin and Zang, Yuhang and Bu, Jiazi and Zhou, Yujie and Xin, Yi and He, Junjun and Wang, Chunyu and Lu, Qinglin and Jin, Cheng and others},
  journal={arXiv preprint arXiv:2510.18701},
  year={2025}
}

@article{wang2025unified,
  title={Unified reward model for multimodal understanding and generation},
  author={Wang, Yibin and Zang, Yuhang and Li, Hao and Jin, Cheng and Wang, Jiaqi},
  journal={arXiv preprint arXiv:2503.05236},
  year={2025}
}

@article{wang2025unified-think,
  title={Unified multimodal chain-of-thought reward model through reinforcement fine-tuning},
  author={Wang, Yibin and Li, Zhimin and Zang, Yuhang and Wang, Chunyu and Lu, Qinglin and Jin, Cheng and Wang, Jiaqi},
  journal={arXiv preprint arXiv:2505.03318},
  year={2025}
}

@article{xu2023imagereward,
  title={Imagereward: Learning and evaluating human preferences for text-to-image generation},
  author={Xu, Jiazheng and Liu, Xiao and Wu, Yuchen and Tong, Yuxuan and Li, Qinkai and Ding, Ming and Tang, Jie and Dong, Yuxiao},
  journal={NeurIPS},
  volume={36},
  pages={15903--15935},
  year={2023}
}

@article{kirstain2023pick,
  title={Pick-a-pic: An open dataset of user preferences for text-to-image generation},
  author={Kirstain, Yuval and Polyak, Adam and Singer, Uriel and Matiana, Shahbuland and Penna, Joe and Levy, Omer},
  journal={NeurIPS},
  volume={36},
  pages={36652--36663},
  year={2023}
}

@article{wang2023enhancing,
  title={Enhancing object coherence in layout-to-image synthesis},
  author={Wang, Yibin and Zhou, Changhai and Xu, Honghui},
  journal={arXiv preprint arXiv:2311.10522},
  year={2023}
}

@article{wang2024magicface,
  title={Magicface: Training-free universal-style human image customized synthesis},
  author={Wang, Yibin and Zhang, Weizhong and Jin, Cheng},
  journal={arXiv preprint arXiv:2408.07433},
  year={2024}
}

@inproceedings{wang2025complexbench,
  title={Complexbench-edit: Benchmarking complex instruction-driven image editing via compositional dependencies},
  author={Wang, Chenglin and Zhou, Yucheng and Wang, Qianning and Wang, Zhe and Zhang, Kai},
  booktitle={Proceedings of the 33rd ACM International Conference on Multimedia},
  pages={13391--13397},
  year={2025}
}

@article{sun2025t2i,
  title={T2i-reasonbench: Benchmarking reasoning-informed text-to-image generation},
  author={Sun, Kaiyue and Fang, Rongyao and Duan, Chengqi and Liu, Xian and Liu, Xihui},
  journal={arXiv preprint arXiv:2508.17472},
  year={2025}
}

@article{li2025easier,
  title={Easier Painting Than Thinking: Can Text-to-Image Models Set the Stage, but Not Direct the Play?},
  author={Li, Ouxiang and Wang, Yuan and Hu, Xinting and Huang, Huijuan and Chen, Rui and Ou, Jiarong and Tao, Xin and Wan, Pengfei and Qi, Xiaojuan and Feng, Fuli},
  journal={arXiv preprint arXiv:2509.03516},
  year={2025}
}
}

\clearpage
\appendix
\hypersetup{pageanchor=false}
\setcounter{page}{1}
\hypersetup{pageanchor=true}

\section*{\centering Supplementary Material}


\definecolor{navy_blue}{RGB}{0, 47, 167}
\definecolor{richCrimson}{RGB}{153, 0, 51}
\definecolor{red_1}{RGB}{241, 185, 184}
\definecolor{blue_1}{RGB}{201, 224, 251}

\section{Detailed Explanation of Table~1 in the Main Paper}
As shown in Tab.~1, existing benchmarks (e.g., RISEBench and KRISBench) are largely \emph{text-major} in evaluation. In contrast, UniREditBench provides reference images for all instances and adopts a \emph{dual-reference} (text+image) protocol, improving evaluation validity and reliability. Moreover, UniREditBench covers a broader range of real-world and game-world scenarios. In the game-world setting, we specify explicit rules and introduce a paradigm in which models perform image editing to execute actions in game environments while adhering to these rules. We further include an intra-game multi-task setting, where each game contains multiple tasks, enabling more comprehensive evaluation.

\section{Weight Selection Strategy}
We determine the weighting scheme by computing the average Pearson correlation coefficient ($r$) between human judgments and GPT-4.1 scores under different weighting settings. As shown in Tab.~\ref{tab:pearson}, the weighting configuration $(0.50, 0.30, 0.20)$ achieves the highest correlation.

\section{Data Filtering}
We design a comprehensive, multi-stage pipeline that performs data filtering, i.e., \textit{instruction de-duplication}, \textit{quality filtering}, and \textit{human inspection}, to remove redundancy and low-quality data. Detailed elaboration of the filtering pipeline is provided below.

\subsection{Instruction De-duplication}
In the first stage of constructing the real-world scenarios, text prompts for image editing are sampled using Gemini-2.5-Pro, which may introduce duplicate entries. We remove redundancy along two aspects: exact matches and semantic similarity.

\begin{itemize}
    \item[\textbullet] \textbf{Exact-Match Deduplication}: We first normalize the \textit{original image description} by converting it to lowercase and removing punctuation. Afterward, we extract the set of words from the normalized text. If two samples contain identical word sets, they are considered duplicates. These duplicate samples are then filtered out to ensure data diversity.
    \item[\textbullet] \textbf{Semantic-Similarity Deduplication}: We use a sentence-transformers mo-\\del to extract sentence embeddings for both the \textit{original image description} and \textit{edit instruction}. We then compute the pairwise similarity between these embeddings. If the similarity score exceeds a threshold of 0.7 for either the description or the instruction, the samples are deemed semantically redundant and are filtered out to enhance dataset diversity.
\end{itemize}

These complementary exact-match and semantic filters improve dataset diversity by eliminating both literal and paraphrastic duplicates.

\begin{table}[t]
\centering
\caption{Average Pearson Correlation Coefficient (r) across different score weighting settings. $\alpha_1$=Instruction Following, $\alpha_2$=Visual Consistency, $\alpha_3$=Visual Quality}
\resizebox{0.98\columnwidth}{!}{%
\begin{tabular}{lcccccc}
\toprule
($\alpha_1, \alpha_2, \alpha_3$) & (0.33,0.33,0.33) & (0.50,0.10,0.20) & (0.50,0.50,0.20)  & (0.30,0.30,0.20) & (0.70,0.30,0.20) & (0.50,0.30,0.20) \\
\midrule
score$\uparrow$   & 0.8357 & 0.8353 & 0.8399  & 0.8383 & 0.8405 & \textbf{0.8408} \\
\bottomrule
\end{tabular}%
}
\label{tab:pearson}
\end{table}

\begin{figure}[t]
    \centering
    \includegraphics[width=\linewidth]{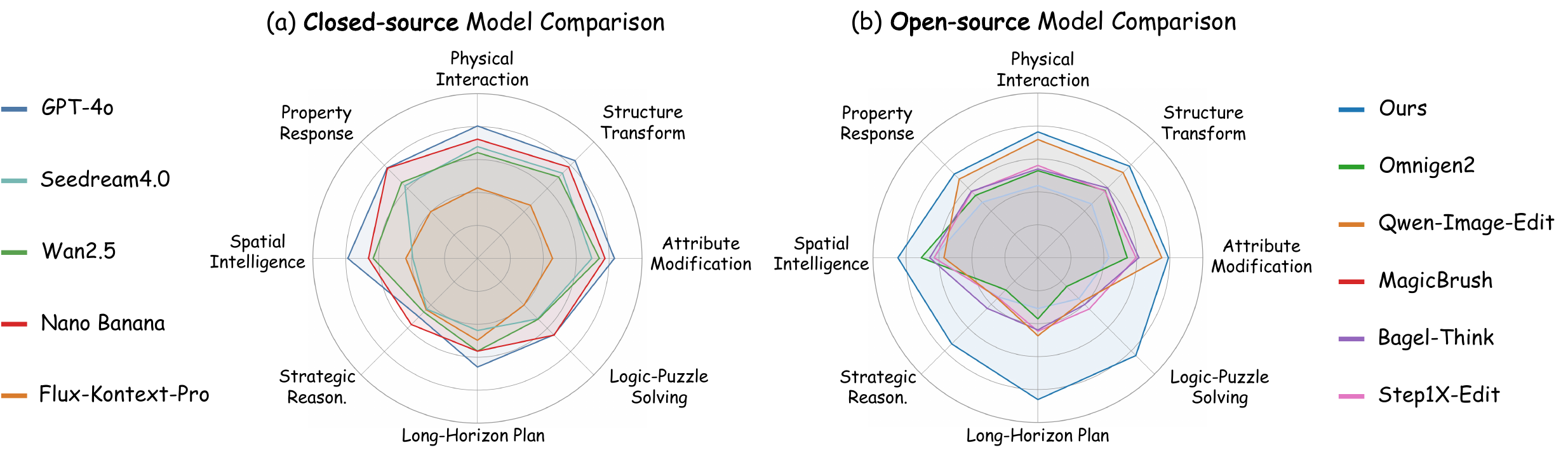} 
    \caption{Benchmarking result visualization. (a) Closed-source model comparison; (b) Open-source model comparison.}
    \label{fig:radar}
\end{figure}

\begin{table}[t]
\centering
\caption{\textbf{Quantitative comparisons on KRISBench}. \textit{GPT-4o} is used as the evaluator. Best scores are in \textbf{bold} while second-best is \underline{underlined}.
}
\scriptsize
\label{tab:result_kris}

\resizebox{\linewidth}{!}{%
\setlength{\tabcolsep}{2pt}
\begin{tabular}{l|ccccccc|cc}
\toprule
\multirow{2}{*}{Model} 
 & Attribute & Spatial & Temporal  & Social & Natural  & Logical & Instruction & \textbf{Overall}\\
 & Perception & Perception & Prediction & Science & Science & Reasoning & Decompose & \textbf{Score} & \\
\midrule

\multicolumn{10}{c}{\textbf{Closed-source Models}} 
\\
\midrule
Doubao & \underline{70.92} & 59.17 & 40.58  & 65.50 & \underline{61.19} & 47.75 & 60.58 & 60.70 \\
Step 3$\phi$ vision & 69.67 & 61.08 & 63.25  & 66.88 & 60.88 & 49.06 & 54.92 & 61.43 \\
\raisebox{-0.2em}{\includegraphics[height=0.8em]{images/second.png}}Gemini-2.0 & 66.33 & \underline{63.33} & \underline{63.92}  & \underline{68.19} & 56.94 & \underline{54.13} & \underline{71.67} & \underline{62.41} \\
\raisebox{-0.2em}{\includegraphics[height=0.8em]{images/winner.png}}GPT-4o  & \textbf{83.17} & \textbf{79.08} & \textbf{68.25} & \textbf{85.50} & \textbf{80.06} & \textbf{71.56} & \textbf{85.08} & \textbf{80.09} \\
\midrule 
\multicolumn{10}{c}{\textbf{Open-source Models}} \\
\midrule
MagicBrush &  53.92 &  39.58 & - & 42.94 & 38.06 &  30.00 & 23.08 & 37.15 \\
AnyEdit & 47.67 & 45.17 & - & 38.56 & 42.94 & 36.56 & 26.92 & 38.55 \\
Emu2 & 51.50 &  48.83 & 22.17 &  34.69 &  38.44 & 24.81 & 45.00 &  39.70 \\
Step1X-Edit & 55.50 & 51.75 & - & 44.69 & 49.06 & 40.88 & 22.75 & 43.29 \\
\raisebox{-0.2em}{\includegraphics[height=0.8em]{images/second.png}}Bagel-Think   & \underline{69.27} & \underline{67.58} & - & \underline{65.00} & \underline{62.11} & \underline{47.33} & \underline{49.22} & \underline{60.77} \\
\hline
\rowcolor[HTML]{E2F4E3}
\raisebox{-0.2em}{\includegraphics[height=0.8em]{images/winner.png}}\textbf{UniREdit-Bagel (Ours)}          & \textbf{71.75} & \textbf{71.00} & - & \textbf{69.20} & \textbf{65.99} & \textbf{59.91} & \textbf{51.55} & \textbf{65.45} \\
\bottomrule
\end{tabular}
}
\end{table}

\subsection{Quality Filtering}
To ensure the quality of both the generated text and images, we evaluate and filter them across six key dimensions: text hallucination, instruction adherence, content preservation, visual quality, image hallucination, and CoT quality. Scores for each dimension are assigned by Gemini-2.5-Pro on a 1-5 scale. Only samples that achieve the maximum score across all dimensions are retained.

\begin{itemize}
    \item[\textbullet] \textbf{Text Hallucination}: We evaluate the textual reference for hallucinated content, defined as entities or visual effects that are not mentioned in the instruction or that cannot be plausibly induced by the given instruction.
    \item[\textbullet] \textbf{Instruction Following}: We compare the edited image with the textual reference to assess whether the generated visual changes accurately reflect the specified effects. Samples that demonstrate poor adherence to the instructions and textual reference are discarded.
    \item[\textbullet] \textbf{Content Preservation}: We assess whether regions unrelated to the edit instruction, such as the background, remain consistent between the original and edited images, ensuring stability in unaffected areas.
    \item[\textbullet] \textbf{Visual Quality}: We assess whether the generated images meet fundamental quality standards, specifically by ensuring they are free from artifacts or degradation.
    \item[\textbullet] \textbf{Image Hallucination}: We examine the edited images for any unintended additions or alterations beyond the specified textual reference, such as the appearance of additional objects.
    \item[\textbullet] \textbf{CoT Quality}: We evaluate the correctness of the chain-of-thought (CoT) reasoning text, focusing on whether the analysis of the original image and instruction is logical and sound.

\end{itemize}

\subsection{Human Inspection}
In addition to our automated filtering pipeline, we perform a final manual check of each data instance. To facilitate this, we developed two web-based interfaces and enlisted eight expert annotators to carry out a two-stage filtering process:

\begin{itemize}
    \item[\textbullet] \textbf{Initial Filtering}: Annotators remove samples with extremely erroneous textual references or substandard generated images.
    \item[\textbullet] \textbf{Manual Correction}: Annotators refine the textual reference effect when it is only slightly incorrect, ensuring better alignment and accuracy.
\end{itemize}

\begin{figure*}[t]
    \centering
    \includegraphics[width=0.8\linewidth]{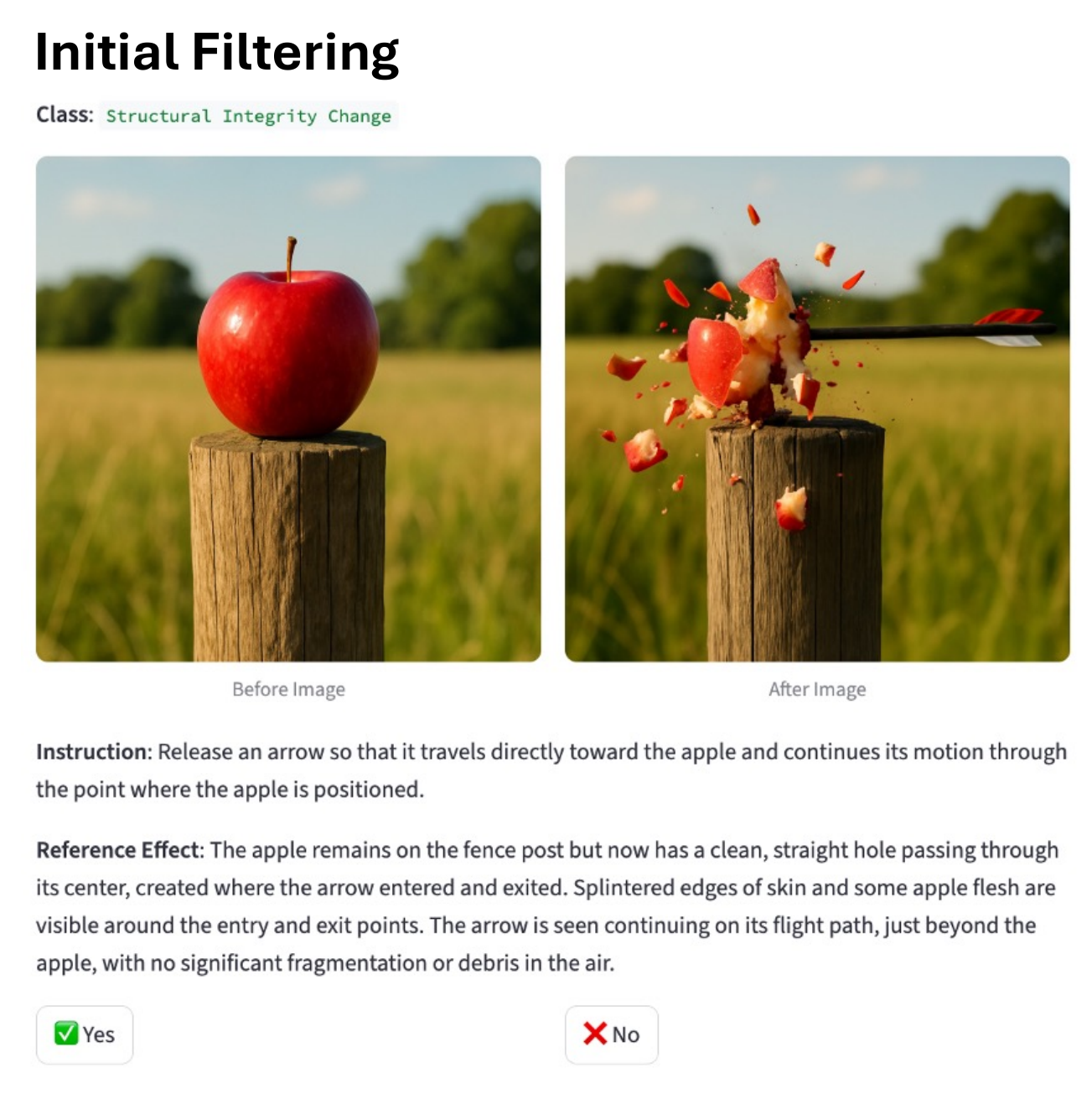} 
    \caption{Web interface of the initial filtering stage.}
    \label{fig:initial_filtering}
    \vspace{-0.5cm}
\end{figure*}

\begin{figure*}[t]
    \centering
    \includegraphics[width=0.8\linewidth]{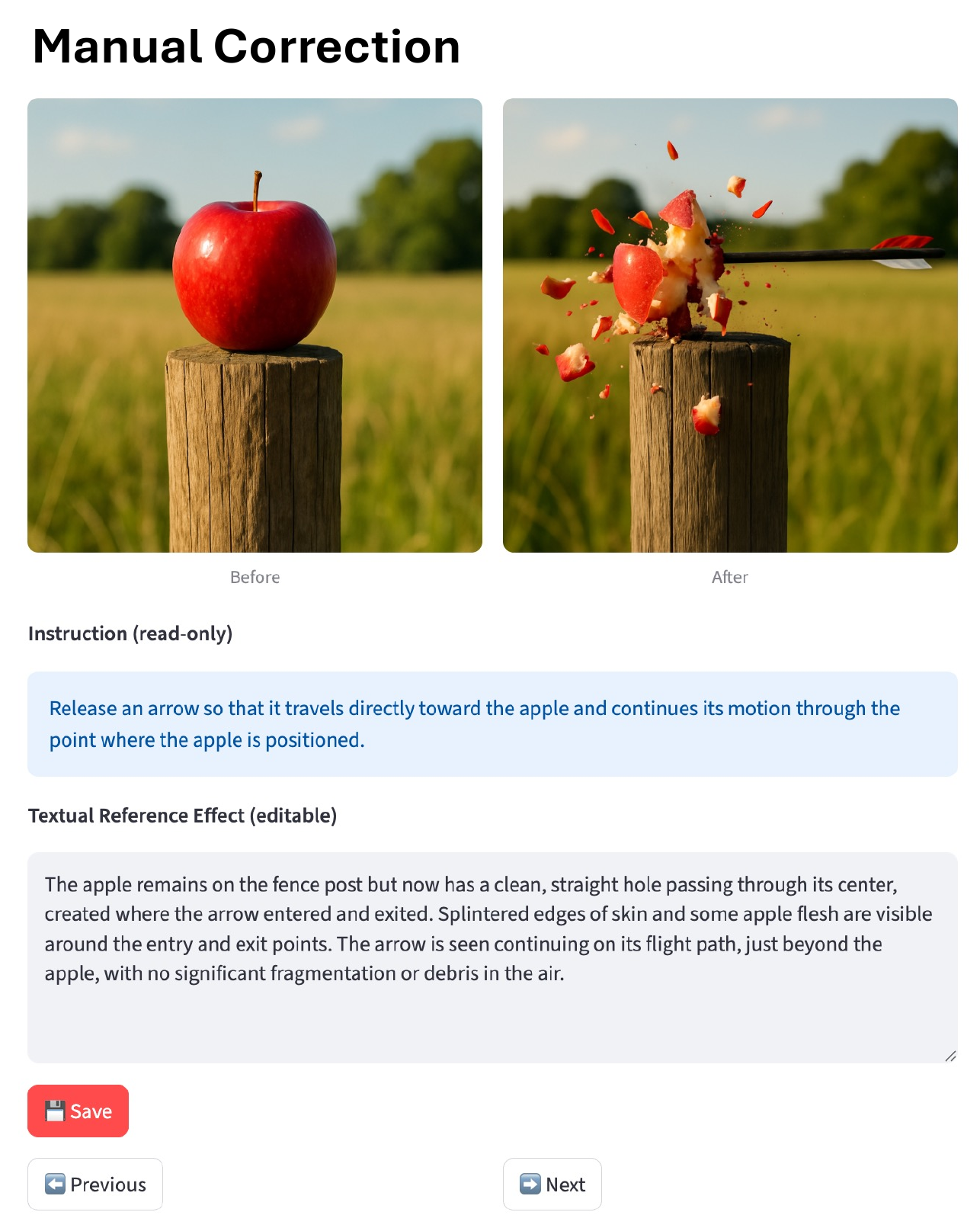} 
    \caption{Web interface of the manual correction stage.}
    \label{fig:manual_correction}
    \vspace{-0.5cm}
\end{figure*}

Two web interfaces are shown in Figs.~\ref{fig:initial_filtering} and ~\ref{fig:manual_correction}.

\section{More Quantitative Results}
We provide additional quantitative comparisons of out-of-distribution performance on KRISBench in Tab.~\ref{tab:result_kris}. Among open-source models, UniREdit-Bagel, fine-tuned on UniREdit-Data-100K, achieves the best performance and even outperforms several closed-source models, including Gemini-2.0 and Doubao.

\section{Detailed Benchmarking Results}
We provide detailed benchmarking results on our UniREditBench for each category in Tab.~\ref{tab:detailed_results}.

\section{Open-source VLM Evaluation}
We provide benchmarking results on UniREditBench using Qwen3VL-32B-Instr-\\uct as the evaluator for each category in Tab.~\ref{tab:detailed_result_qwen}. The results show that Qwen3VL-32B-Instruct yields scoring trends consistent with those of GPT-4.1 (Tab.~\ref{tab:detailed_results}).

\section{Transferability of Game-world Data}
We conduct an ablation study to examine the effect of game-world data when evaluating on real-world scenarios (Tab.~\ref{tab:game_ablate}). Incorporating game-world data consistently improves performance across all four categories. This indicates a positive cross-domain transfer from game-world tasks to real-world scenarios.

\begin{table*}[t]
    \centering
    \caption{Detailed in-domain quantitative comparisons on UniREditBench. \textit{GPT-4.1} is used as the evaluator. Best scores are in \textbf{bold}.}
    \tiny
    \label{tab:detailed_results}
    \resizebox{\linewidth}{!}{%
    \setlength{\tabcolsep}{2pt}
    \begin{tabular}{l|ccccccccc|ccccccccc|c}
        \toprule
        \multirow{8}{*}{Model} 
        & \multicolumn{9}{c|}{\textbf{Real World Scenario}} 
        & \multicolumn{9}{c|}{\textbf{Game World Scenario}} 
        & \multirow{8}{*}{Overall} \\
        \cmidrule(lr){2-10}\cmidrule(lr){11-19}
        & \multicolumn{2}{c}{\makecell{Attribute\\Modification}} 
        & \multicolumn{2}{c}{\makecell{Structure\\Transform}} 
        & \multicolumn{3}{c}{\makecell{Physical\\Interaction}} 
        & \multicolumn{2}{c|}{\makecell{Property\\Response}} 
        & \multicolumn{1}{c}{\makecell{Spatial\\Intelligence}} 
        & \multicolumn{3}{c}{\makecell{Strategic\\Reason}} 
        & \multicolumn{3}{c}{\makecell{Logic\\Puzzle Solving}} 
        & \multicolumn{2}{c|}{\makecell{Long-Horizon\\Plan}} 
        & \\
        \cmidrule(lr){2-3}\cmidrule(lr){4-5}\cmidrule(lr){6-8}\cmidrule(lr){9-10}
        \cmidrule(lr){11-11}\cmidrule(lr){12-14}\cmidrule(lr){15-17}\cmidrule(lr){18-19}
        & \makecell{Viewpoint\\Transformation} 
        & \makecell{Material\\Modification} 
        & \makecell{Pose\\Adjustment} 
        & \makecell{Temporal\\Evolution} 
        & \makecell{Structural\\Integrity\\Change} 
        & \makecell{Motion\\State\\Change} 
        & \makecell{Spatial\\Arrangement} 
        & \makecell{Mechanical\\Reaction} 
        & \makecell{Medium\\Interaction} 
        & \makecell{3D\\Reconstruction} 
        & \makecell{Space\\Invader} 
        & Jewel2 
        & Pacman 
        & \makecell{Word\\Search} 
        & Tic-tac-toe 
        & Sudoku 
        & Maze 
        & Sokoban 
        & \\
        \midrule
        \multicolumn{20}{c}{\textbf{Closed-source Models}} \\
        \midrule
        FLUX-Kontext-Pro 
        & 37.55 & 57.15 & 56.79 & 37.52 & 41.46 & 44.62 & 47.03 & 40.56 & 42.32 
        & 49.12 & 53.18 & 38.12 & 54.04 & 32.45 & \textbf{61.23} & 27.80 & 48.45 & \textbf{53.87} & 45.77\\
        Seedream4.0 
        & 64.77 & 74.32 & 80.19 & 66.07 & 59.48 & 64.68 & 79.48 & 61.58 & 63.21 
        & 39.27 & 46.28 & 43.00 & 41.34 & 48.55 & 38.43 & 68.75 & 54.15 & 33.43 & 57.03\\
        Wan2.5 
        & 73.78 & 75.54 & 79.95 & 59.53 & 62.67 & 64.38 & 63.49 & 60.83 & 64.91 
        & 63.73 & 43.44 & \textbf{56.57} & 41.38 & 58.38 & 58.57 & 47.86 & 64.03 & 45.97 & 61.36\\
        \raisebox{-0.2em}{\includegraphics[height=0.8em]{images/second.png}}Nano Banana 
        & 75.82 & 78.39 & 86.12 & 71.63 & 71.37 & 71.07 & 73.14 & 77.58 & 71.82 
        & 66.74 & \textbf{59.97} & 54.33 & \textbf{54.43} & 64.13 & 39.80 & \textbf{90.81} & 62.50 & 51.15 & 68.26\\
        \raisebox{-0.2em}{\includegraphics[height=0.8em]{images/winner.png}}GPT-4o 
        & \textbf{83.83} & \textbf{81.52} & \textbf{92.18} & \textbf{77.33} & \textbf{75.86} & \textbf{74.98} & \textbf{88.60} & \textbf{78.98} & \textbf{75.81} 
        & \textbf{77.73} & 58.96 & 45.60 & 49.75 & \textbf{64.62} & 58.97 & 67.78 & \textbf{83.73} & 51.49 & \textbf{73.39}\\
        \midrule
        \multicolumn{20}{c}{\textbf{Open-source Models}} \\
        \midrule
        MagicBrush 
        & 36.74 & 51.20 & 47.75 & 44.43 & 44.05 & 47.06 & 37.67 & 45.87 & 47.42 
        & 63.58 & 45.33 & 25.60 & 29.83 & 33.92 & 31.45 & 40.55 & 33.20 & 28.23 & 40.77\\
        Omnigen2 
        & 51.25 & 59.68 & 68.25 & 48.28 & 48.37 & 54.48 & 51.48 & 50.77 & 50.64 
        & 70.28 & 51.33 & 1.38 & 29.80 & 22.87 & 44.97 & 5.10 & 39.57 & 32.93 & 43.41\\
        Bagel 
        & 58.55 & 65.15 & 67.23 & 51.87 & 52.40 & 57.38 & 47.88 & 54.22 & 54.72 
        & 54.25 & 37.43 & 26.02 & 42.18 & 40.47 & 46.13 & 32.90 & 43.68 & 31.70 & 48.01\\
        Lumina-DiMOO 
        & 54.22 & 51.47 & 54.40 & 50.23 & 48.67 & 52.62 & 49.65 & 51.63 & 50.03 
        & 61.23 & 48.23 & 36.29 & 26.35 & 41.73 & 65.07 & 52.47 & 43.57 & 35.57 & 48.54\\
        Step1X-Edit 
        & 52.57 & 66.82 & 62.15 & 50.58 & 53.07 & 58.68 & 49.80 & 54.23 & 55.45 
        & 65.68 & 35.05 & 36.38 & 33.25 & 45.49 & 46.97 & 39.20 & 55.94 & 38.08 & 50.15\\
        Bagel-Think 
        & 59.05 & 63.84 & 63.52 & 55.49 & 54.05 & 55.62 & 48.27 & 55.03 & 56.33 
        & 66.29 & 45.48 & 42.89 & 41.33 & 48.42 & 42.20 & 33.27 & 48.62 & 37.38 & 50.96\\
        DreamOmni2 
        & 61.22 & 65.82 & 67.07 & 49.63 & 49.58 & 54.83 & 53.62 & 54.11 & 53.92 
        & 72.42 & 38.78 & 42.33 & 45.40 & 45.86 & 46.98 & 51.42 & 56.55 & 41.05 & 52.81\\
        UniWorld-V2 
        & 71.45 & 74.47 & 83.19 & 55.55 & 58.53 & 65.47 & 66.95 & 62.20 & 61.17 
        & 49.27 & 34.44 & 42.90 & 42.65 & 49.67 & 34.73 & 28.20 & 71.59 & 35.22 & 54.87\\
        \raisebox{-0.2em}{\includegraphics[height=0.8em]{images/second.png}}Qwen-Image-Edit 
        & 74.60 & \textbf{76.76} & 81.45 & 64.62 & 66.48 & 68.78 & 76.50 & 64.65 & 64.69 
        & 56.73 & 38.42 & 41.02 & 30.45 & 48.52 & 33.60 & 30.93 & 61.38 & 36.21 & 56.52\\
        \hline
        \rowcolor[HTML]{E2F4E3}
        \raisebox{-0.2em}{\includegraphics[height=0.8em]{images/winner.png}}\textbf{UniREdit-Bagel }(Ours) 
        & \textbf{81.08} & 72.39 & \textbf{84.78} & \textbf{70.81} & \textbf{73.85} & \textbf{72.73} & \textbf{83.13} & \textbf{71.27} & \textbf{71.62} 
        & \textbf{84.90} & \textbf{87.77} & \textbf{57.68} & \textbf{73.05} & \textbf{86.80} & \textbf{71.16} & \textbf{93.20} & \textbf{98.30} & \textbf{71.47} & \textbf{78.15}\\
        \bottomrule
    \end{tabular}
    }
\end{table*}

\begin{table*}[t]
    \centering
    \caption{Detailed in-domain quantitative comparisons on UniREditBench. \textit{Qwen3VL-32B-Instruct} is used as the evaluator. Best scores are in \textbf{bold}.}
    \tiny
    \label{tab:detailed_result_qwen}
    \resizebox{\linewidth}{!}{%
    \setlength{\tabcolsep}{2pt}
    \begin{tabular}{l|ccccccccc|ccccccccc|c}
        \toprule
        \multirow{8}{*}{Model} 
        & \multicolumn{9}{c|}{\textbf{Real World Scenario}} 
        & \multicolumn{9}{c|}{\textbf{Game World Scenario}} 
        & \multirow{8}{*}{Overall} \\
        \cmidrule(lr){2-10}\cmidrule(lr){11-19}
        & \multicolumn{2}{c}{\makecell{Attribute\\Modification}} 
        & \multicolumn{2}{c}{\makecell{Structure\\Transform}} 
        & \multicolumn{3}{c}{\makecell{Physical\\Interaction}} 
        & \multicolumn{2}{c|}{\makecell{Property\\Response}} 
        & \multicolumn{1}{c}{\makecell{Spatial\\Intelligence}} 
        & \multicolumn{3}{c}{\makecell{Strategic\\Reason}} 
        & \multicolumn{3}{c}{\makecell{Logic\\Puzzle Solving}} 
        & \multicolumn{2}{c|}{\makecell{Long-Horizon\\Plan}} 
        & \\
        \cmidrule(lr){2-3}\cmidrule(lr){4-5}\cmidrule(lr){6-8}\cmidrule(lr){9-10}
        \cmidrule(lr){11-11}\cmidrule(lr){12-14}\cmidrule(lr){15-17}\cmidrule(lr){18-19}
        & \makecell{Viewpoint\\Transformation} 
        & \makecell{Material\\Modification} 
        & \makecell{Pose\\Adjustment} 
        & \makecell{Temporal\\Evolution} 
        & \makecell{Structural\\Integrity\\Change} 
        & \makecell{Motion\\State\\Change} 
        & \makecell{Spatial\\Arrangement} 
        & \makecell{Mechanical\\Reaction} 
        & \makecell{Medium\\Interaction} 
        & \makecell{3D\\Reconstruction} 
        & \makecell{Space\\Invader} 
        & Jewel2 
        & Pacman 
        & \makecell{Word\\Search} 
        & Tic-tac-toe 
        & Sudoku 
        & Maze 
        & Sokoban 
        & \\
        \midrule
        \multicolumn{20}{c}{\textbf{Closed-source Models}} \\
        \midrule
        FLUX-Kontext-Pro 
        & 56.25 & 62.97 & 61.34 & 51.83 & 49.91 & 52.03 & 50.30 & 51.38 & 52.50 
        & 55.22 & 34.87 & 23.84 & 26.47 & 33.37 & \textbf{58.18} & 39.03 & 40.62 & \textbf{47.80} & 46.99 \\
        Seedream4.0 
        & 61.28 & 76.00 & 81.82 & 62.67 & 55.32 & 57.13 & 82.83 & 58.38 & 60.49 
        & 36.37 & 32.83 & 43.20 & 37.82 & 41.88 & 43.47 & 67.55 & 53.40 & 37.83 & 55.01\\
        Wan2.5 
        & 67.73 & 75.07 & 74.20 & 53.09 & 57.05 & 56.76 & 58.27 & 55.20 & 58.90 
        & 49.62 & 36.48 & 46.88 & 37.83 & 45.56 & 47.54 & 49.53 & 63.20 & 40.61 & 55.11\\
        \raisebox{-0.2em}{\includegraphics[height=0.8em]{images/second.png}}Nano Banana 
        & 67.83 & 77.05 & 81.71 & 64.46 & 65.05 & 62.12 & 71.50 & 68.77 & 65.90 
        & 64.27 & \textbf{44.38} & \textbf{48.07} & \textbf{48.71} & 49.60 & 45.13 & \textbf{90.28} & 59.63 & 47.05 & 62.87\\
        \raisebox{-0.2em}{\includegraphics[height=0.8em]{images/winner.png}}GPT-4o 
        & \textbf{82.07} & \textbf{84.20} & \textbf{89.29} & \textbf{75.13} & \textbf{73.17} & \textbf{67.72} & \textbf{86.85} & \textbf{77.30} & \textbf{73.42} 
        & \textbf{66.22} & 43.08 & 32.91 & 39.77 & \textbf{50.23} & 56.83 & 58.30 & \textbf{66.01} & 32.52 & \textbf{64.10} \\
        \midrule
        \multicolumn{20}{c}{\textbf{Open-source Models}} \\
        \midrule
        MagicBrush 
        & 31.97 & 45.68 & 41.58 & 39.65 & 43.23 & 41.50 & 36.75 & 44.22 & 44.97 
        & 62.45 & 45.30 & 20.86 & 34.40 & 34.42 & 22.25 & 40.52 & 34.25 & 22.07 & 38.12\\
        Omnigen2 
        & 48.20 & 52.03 & 62.68 & 46.02 & 46.27 & 48.00 & 47.87 & 49.52 & 47.98 
        & 47.25 & 44.02 & 0.57 & 28.38 & 25.25 & 33.87 & 3.17 & 44.30 & 29.50 & 39.16\\
        Bagel 
        & 55.05 & 60.33 & 61.12 & 47.17 & 50.00 & 51.20 & 46.97 & 50.65 & 50.18 
        & 47.77 & 30.93 & 21.79 & 37.94 & 38.28 & 39.85 & 36.48 & 44.19 & 35.97 & 44.79 \\
        Step1X-Edit 
        & 49.17 & 63.52 & 56.77 & 47.57 & 50.00 & 52.30 & 48.10 & 51.97 & 53.57 
        & 50.02 & 25.65 & 21.70 & 29.95 & 37.77 & 39.85 & 40.23 & 54.28 & 38.58 & 45.05\\
        Lumina-DiMOO 
        & 50.60 & 46.60 & 50.12 & 46.23 & 49.47 & 49.10 & 47.30 & 50.32 & 50.23 
        & 50.93 & 46.30 & 46.40 & 37.43 & 45.02 & 49.08 & 48.43 & 47.15 & 47.08 & 47.66\\
        Bagel-Think 
        & 54.45 & 60.03 & 57.13 & 49.58 & 52.02 & 50.38 & 47.97 & 52.72 & 52.45 
        & 56.62 & 38.15 & 41.88 & 40.27 & 44.97 & 38.43 & 38.12 & 48.57 & 41.03 & 48.05\\
        DreamOmni2 
        & 55.00 & 61.43 & 61.12 & 47.87 & 47.65 & 52.37 & 49.98 & 51.77 & 51.75 
        & 63.30 & 30.65 & 30.78 & 42.57 & 40.65 & 50.30 & 48.53 & 51.28 & 42.47 & 48.86\\
        UniWorld-V2 
        & 67.36 & 74.38 & 79.26 & 50.09 & 56.51 & 60.24 & 67.94 & 59.33 & 59.18 
        & 35.58 & 33.76 & 22.63 & 28.26 & 40.69 & 24.09 & 26.09 & 69.44 & 28.31 & 49.07 \\
        \raisebox{-0.2em}{\includegraphics[height=0.8em]{images/second.png}}Qwen-Image-Edit 
        & 65.69 & \textbf{74.92} & 76.41 & 57.48 & 61.66 & 59.45 & 73.64 & 58.49 & 60.66 
        & 51.45 & 31.96 & 26.81 & 30.37 & 40.64 & 30.88 & 28.65 & 59.18 & 32.32 & 51.06\\
        \hline
        \rowcolor[HTML]{E2F4E3}
        \raisebox{-0.2em}{\includegraphics[height=0.8em]{images/winner.png}}\textbf{UniREdit-Bagel }(Ours) 
        & \textbf{78.72} & 73.45 & \textbf{80.80} & \textbf{68.48} & \textbf{73.22} & \textbf{65.88} & \textbf{83.58} & \textbf{69.95} & \textbf{67.48} 
        & \textbf{76.03} & \textbf{79.00} & \textbf{54.88} & \textbf{63.22} & \textbf{85.25} & \textbf{68.65} & \textbf{94.57} & \textbf{94.97} & \textbf{76.88} & \textbf{75.29}\\
        \bottomrule
    \end{tabular}
    }
\end{table*}

\begin{table}[t]
\centering
\caption{Ablation on game-world data in real-world scenarios.}
\tiny
\setlength{\tabcolsep}{4pt}
\resizebox{0.98\linewidth}{!}{%
\begin{tabular}{l|ccccc}
\toprule
\multirow{2}{*}{Model} &
Attribute & Structure & Physical & Property & \multirow{2}{*}{Avg.} \\
& Modification & Transform & Interaction & Response & \\
\midrule
UniREdit-Bagel (w/o game data)   & 74.78 & 76.74 & 74.22 & 69.50 & 73.81 \\
UniREdit-Bagel (with game data)  & 76.73 & 77.80 & 76.57 & 71.44 & \textbf{75.74} \\
\bottomrule
\end{tabular}
}
\label{tab:game_ablate}
\end{table}

\begin{table}[h]
\centering
\caption{Evaluation of CoT quality and alignment between edits and CoT.}
\resizebox{0.7\linewidth}{!}{%
\begin{tabular}{lc|c}
\toprule
 & \ \ CoT with causal reasoning \ \  & \ \ Edited images aligned with CoT \ \ \\
\midrule
Score (0-1) & 0.99 & 0.84 \\
\bottomrule
\end{tabular}
}
\label{tab:cot_alignment}
\end{table}

\begin{figure}[t]
    \centering
    \includegraphics[width=0.6\linewidth]{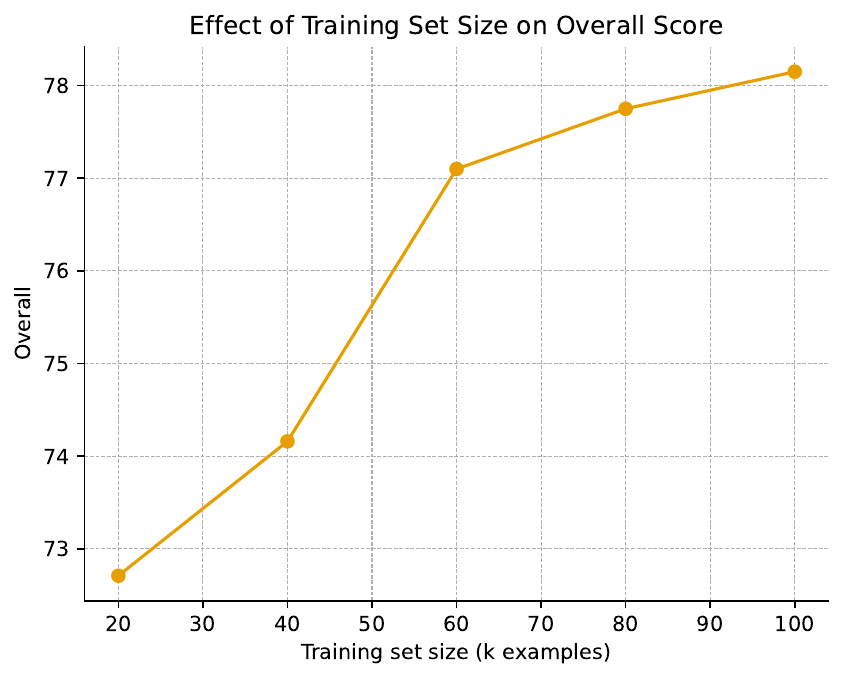}
    \caption{Overall performance on UniREditBench of models trained on different sizes of UniREdit-Data.}
    \label{fig:data_scale_up_effect}
    \vspace{-0.3cm}
\end{figure}

\section{Role of Chain-of-Thought Reasoning}
We further investigate the role of CoT in guiding image editing. Specifically, we use Gemini-3-Pro as a judge to assess two binary criteria: (i) whether the CoT contains explicit causal reasoning, and (ii) whether the final edited image is aligned with the CoT. As shown in Tab.~\ref{tab:cot_alignment}, CoTs provide clear causal chains to steer the editing process toward the intended outcome.

\section{Effect of Training Set Size on Overall Score}
We compare the overall performance on UniREditBench of models trained with different amounts of UniREdit-Data, as illustrated in Fig.~\ref{fig:data_scale_up_effect}. The results show consistent improvements as the training set size increases from 20K to 100K samples.

\section{More Qualitative Comparison Results}
\begin{figure*}[t] 
    \centering
    \includegraphics[width=0.9\linewidth]{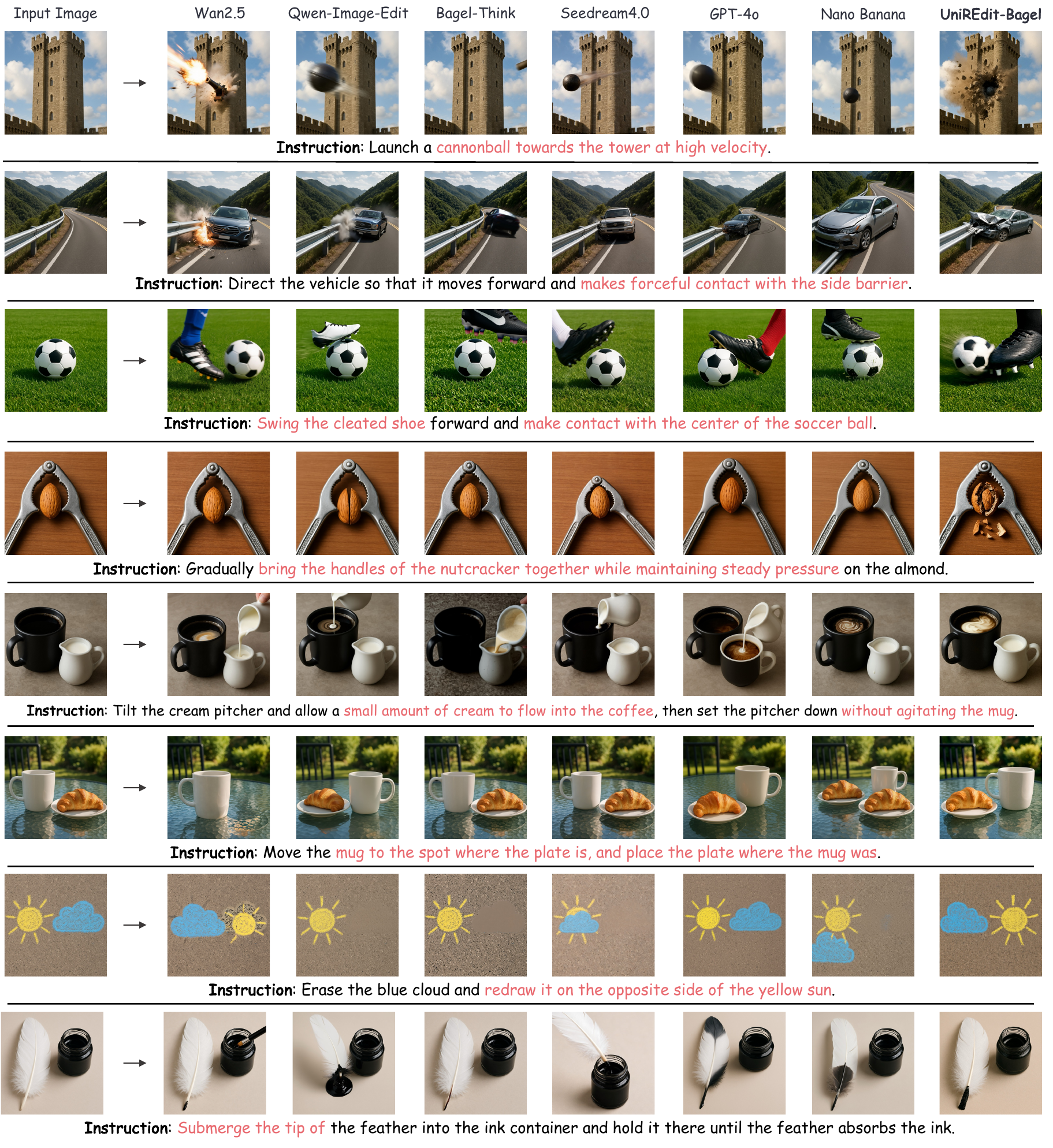} 
    \caption{Qualitative editing result comparison on UniREditBench. Our UniREdit-Bagel demonstrates significant superiority in both instruction following and visual quality compared with state-of-the-art closed-source and open-source models.}
    \label{fig:append_uniredit_qual1}
    \vspace{-0.5cm}
\end{figure*}

\begin{figure*}[t]
    \centering
    \includegraphics[width=0.95\linewidth]{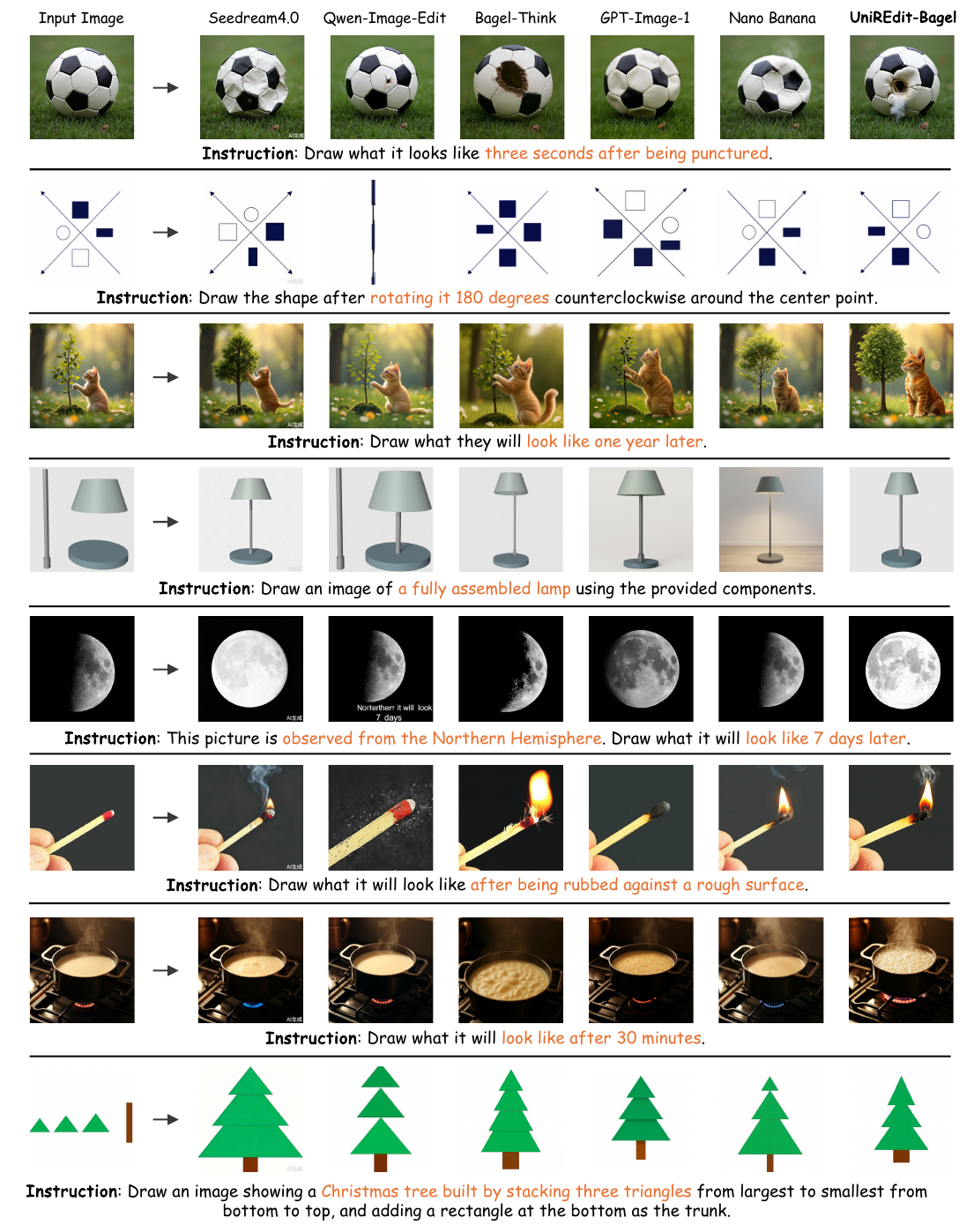} 
    \caption{Qualitative editing result comparison on RISEBench. Our UniREdit-Bagel demonstrates significant superiority in both instruction following and
    visual quality compared with state-of-the-art closed-source and open-source models.}
    \label{fig:rise_qual}
    \vspace{-0.5cm}
\end{figure*}

We provide additional qualitative comparisons on UniREditBench in Figs.~\ref{fig:append_uniredit_qual1} and~\ref{fig:append_uniredit_qual2}, and comparisons on RISEBench in Fig.~\ref{fig:rise_qual}.

\section{Specific Model Failure Modes in UniREditBench}
\begin{figure*}[t]
    \centering
    \includegraphics[width=0.9\linewidth]{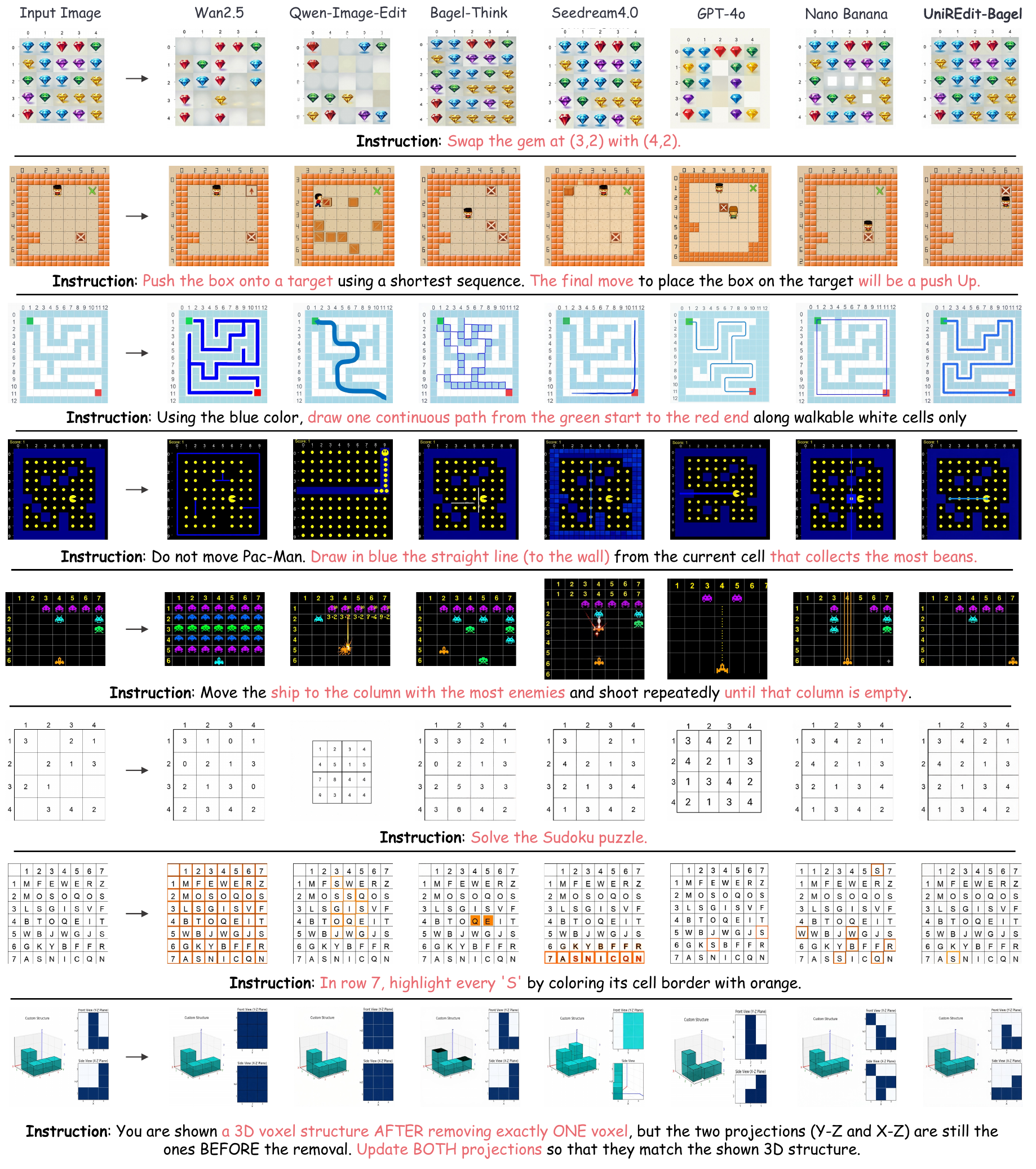} 
    \caption{Qualitative editing result comparison on UniREditBench. Our UniREdit-Bagel demonstrates significant superiority in both instruction following and visual quality compared with state-of-the-art closed-source and open-source models.}
    \label{fig:append_uniredit_qual2}
    \vspace{-0.5cm}
\end{figure*}

As illustrated in Rows~1--2 of Fig.~\ref{fig:append_uniredit_qual2}, UniREditBench exposes a representative failure mode, which we term \textbf{Logical Hallucination}. Specifically, multiple models violate object-permanence rules by deleting non-target gems in Jewel2 (Wan2.5), or violate conservation laws by duplicating boxes in Sokoban (Bagel-Think). These failures underscore the unique challenge posed by UniREditBench: image editing models must not only produce visually plausible outputs, but also adhere to strict logical rules and maintain state consistency in structured environments. This reveals a critical limitation in current models’ perception-and-reasoning capabilities.

\section{Game-world Rules for UniREditBench}
\noindent We summarize the explicit rules for each game category as follows.

\noindent\textbf{3DReconstruction.}
$3\times3\times3$ grid; coordinates $(x,y,z)\in\{1,2,3\}$; at most one voxel per cell; all voxels must be face-connected; new voxels can only be placed adjacent to existing ones. YZ view looks along $-X$ (y horizontal, z vertical); XZ view looks along $+Y$ (x horizontal, z vertical); a projection cell is 1 if any voxel lies on that line.

\noindent\textbf{Jewel2.}
(1) The board uses only A--E gems. (2) Swap two orthogonally adjacent cells. (3) Elimination occurs only when forming a horizontal/vertical run of $\geq3$ identical gems. (4) Diagonals do not count. (5) Eliminated cells become empty and remain empty (no gravity/new gems). (6) No cascades.

\noindent\textbf{Maze.}
Light-blue cells are walls; green is the start; red is the end; white cells are walkable. Find and draw the single valid path from start to end.

\noindent\textbf{Pacman.}
(1) Eating a bean adds 1 point. (2) Pacman cannot pass through walls (movement into blue tiles is blocked). (3) Blue tiles represent walls.

\noindent\textbf{Sokoban.}
Coordinates are $(x,y)$ with $x$ as row and $y$ as column. The player moves up/down/left/right into an empty square and can push one box into an adjacent empty square. Boxes cannot be pulled or pushed through walls or other boxes. The goal is to move all boxes onto target squares.

\noindent\textbf{Space Invader.}
(1) The grid and enemies are fixed (enemies do not move). (2) The ship is below the grid and can move to any column. (3) Shooting a column destroys only the lowermost enemy in that column. (4) Scores: purple=30, blue=20, green=10.

\noindent\textbf{Sudoku.}
\begin{itemize}
    \item[\textbullet] Variant A: Four independent mini Sudokus arranged in a $2\times2$ grid. Each mini board is filled with numbers $1$--$2$ with no duplicates in any row, column, or sub-box within that mini board.
    \item[\textbullet] Variant B: Fill the grid with numbers $1$--$3$ with no duplicates in any row, column, or sub-box.
    \item[\textbullet] Variant C: Fill the grid with numbers $1$--$4$ with no duplicates in any row, column, or sub-box.
\end{itemize}

\noindent\textbf{Tic-tac-toe.}
Tic-tac-toe on a $3\times3$ grid. O moves first. Determine the side to move by counts: if O=X, then O to move; otherwise X to move. Win by 3 in a row (row/column/diagonal); draw if the board is full. Rendering/editing: O=red block, X=blue block. Change only the specified cell, keep the grid and coordinate labels, and output only the updated board image. For optimal move selection, use: (1) immediate win; (2) block opponent's immediate win; (3) create a double threat; (4) block opponent's double threat; (5) otherwise choose the first empty cell in scan order $(0,0)\rightarrow(2,2)$.

\noindent\textbf{Word Search.}
\begin{itemize}
    \item[\textbullet] Variant A: The grid contains uppercase letters. Highlight all cells containing a target letter within a designated row/column by coloring their borders.
    \item[\textbullet] Variant B: Hidden words are placed horizontally or vertically. Highlight the cells forming a target word by coloring the word's outer border.
\end{itemize}

\section{Ethical statement}
In this work, we affirm our commitment to ethical research practices and responsible innovation. To the best of our knowledge, this study does not involve any data, methodologies, or applications that raise ethical concerns. All experiments and analyses were conducted in compliance with established ethical guidelines, ensuring the integrity and transparency of our research process.

\clearpage

\clearpage

\clearpage

\clearpage

\begin{figure*}[t]
    \centering
    \includegraphics[width=0.8\linewidth]{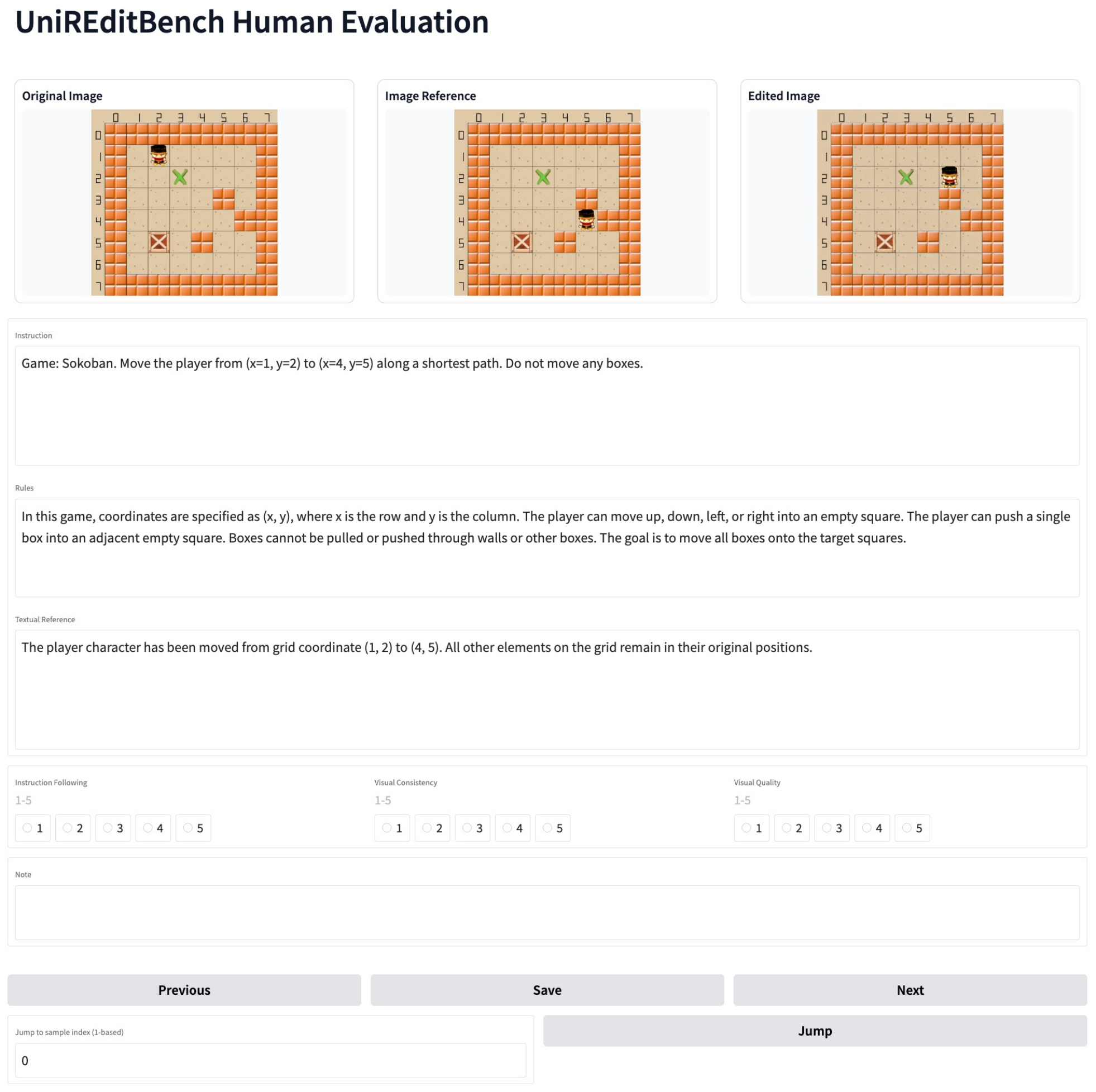} 
    \caption{Web interface for human evaluation.}
    \label{fig:human_eval_platform}
    \vspace{-0.5cm}
\end{figure*}

\end{document}